\documentclass[lettersize,journal]{IEEEtran}

\usepackage{cite}
\usepackage{graphicx}
\usepackage{multirow}
\usepackage{amssymb}
\usepackage[pagebackref=false,breaklinks=true,letterpaper=true,colorlinks,bookmarks=false]{hyperref}
\usepackage[caption=false,font=normalsize,labelfont=sf,textfont=sf]{subfig}
\usepackage{tabularx}
\usepackage{booktabs,chemformula}
\usepackage{amsfonts,amssymb}
\usepackage{pdfpages}
\usepackage{soul}
\usepackage{amsmath}
\usepackage[linewidth=1pt]{mdframed}
\usepackage{multirow}
\usepackage{microtype}
\usepackage{physics}
\usepackage{multicol}
\usepackage{tabularx}
\usepackage{titlesec}

\usepackage{graphicx}
\usepackage[export]{adjustbox}
\usepackage{xcolor}
\usepackage{float}

\newcommand{\RCC}[1]{\textcolor{black}{ #1}}
\newcommand{\RCCC}[1]{\textcolor{black}{ #1}}
\usepackage{threeparttable}

\usepackage{orcidlink}
\titlespacing*{\subsubsection}{0pt}{3.25ex plus 1ex minus .2ex}{1.5ex plus .2ex}

\begin{document}

\title{BP-SGCN: Behavioral Pseudo-Label Informed Sparse Graph Convolution Network for Pedestrian and Heterogeneous Trajectory Prediction 
}



\author{Ruochen Li$^{\orcidlink{0000-0001-8966-9613}}$, Stamos Katsigiannis$^{\orcidlink{0000-0001-9190-0941}}$,~\IEEEmembership{Member,~IEEE}, Tae-Kyun Kim$^{\orcidlink{0000-0002-7587-6053}}$, Hubert P. H. Shum$^{\orcidlink{0000-0001-5651-6039}\dag}$,~\IEEEmembership{Senior Member,~IEEE}

\thanks{R. Li, S. Katasigiannis and H. P. H. Shum are with Durham University, UK.
        (e-mail: \{ruochen.li, stamos.katsigiannis, hubert.shum\}@durham.ac.uk).}%
\thanks{T. K. Kim is with KAIST, Korea.
        (e-mail: kimtaekyun@kaist.ac.kr).}%
\thanks{$^{\dag}$ Corresponding author: H. P. H. Shum}
}


\markboth{Journal of \LaTeX\ Class Files,~Vol.~14, No.~8, August~2021}%
{Shell \MakeLowercase{\textit{et al.}}: A Sample Article Using IEEEtran.cls for IEEE Journals}

\maketitle

\begin{abstract}
Trajectory prediction allows better decision-making in applications of autonomous vehicles or surveillance by predicting the short-term future movement of traffic agents. It is classified into pedestrian or heterogeneous trajectory prediction. The former exploits the relatively consistent behavior of pedestrians, but is limited in real-world scenarios with heterogeneous traffic agents such as cyclists and vehicles. The latter typically relies on extra class label information to distinguish the heterogeneous agents, but such labels are costly to annotate and cannot be generalized to represent different behaviors within the same class of agents. In this work, we introduce the behavioral pseudo-labels that effectively capture the behavior distributions of pedestrians and heterogeneous agents solely based on their motion features, significantly improving the accuracy of trajectory prediction. To implement the framework, we propose the Behavioral Pseudo-Label Informed Sparse Graph Convolution Network (BP-SGCN) that learns pseudo-labels and informs to a trajectory predictor. For optimization, we propose a cascaded training scheme, in which we first learn the pseudo-labels in an unsupervised manner, and then perform end-to-end fine-tuning on the labels in the direction of increasing the trajectory prediction accuracy. 
Experiments show that our pseudo-labels effectively model different behavior clusters and improve trajectory prediction. Our proposed BP-SGCN outperforms existing methods using both pedestrian (ETH/UCY, pedestrian-only SDD) and heterogeneous agent datasets (SDD, Argoverse 1).

\end{abstract}

\section{Introduction}
\label{sec1} 
\IEEEPARstart{P}{redicting} the future movement of traffic agents, known as trajectory prediction, is crucial for safe and efficient decision-making in applications such as autonomous vehicles \cite{luo2018porca}.
\RCCC{Thanks to reliable data-driven \cite{yang2024VibrationControl} object tracking methods \cite{joseph2015yolo}, accurate geometric trajectories can be extracted from videos, serving as a more representative feature set for modeling. Graph Convolutional Networks (GCNs) \cite{kipf2016GCN} have shown exceptional performance across diverse fields due to their adeptness at capturing spatial relationships \cite{Li2024GuiestEditorial, zheng2022towardGraphSelf, Li2024Heterophilous, wu2023GraphIncor, luo2022DirectedNetwork}. This enables them to excel in applications ranging from trajectory agent interaction modeling \cite{shi2021sgcn, huang2019stgat, Mohamed2020socialstgcnn, mohamed2022socialimplicit} to human skeleton-based behavior modeling \cite{xu2023DecoupleSqueeze, shu2022anchor-contrastive, xu2022xinvariant, xu2023polymerization, qiao20222ggcn}, highlighting the superior capabilities in handling graph-based data structures.} Similarly, recognizing distinct movement behavior patterns among agents is pivotal to model the temporal dependency \cite{sun2021PCCSNet}. These patterns, when integrated with GCN, further enhance the precision of predictions by accounting for the inherent behavioral tendencies.

\begin{figure}[t]\centering
  \includegraphics[width=0.48\textwidth]{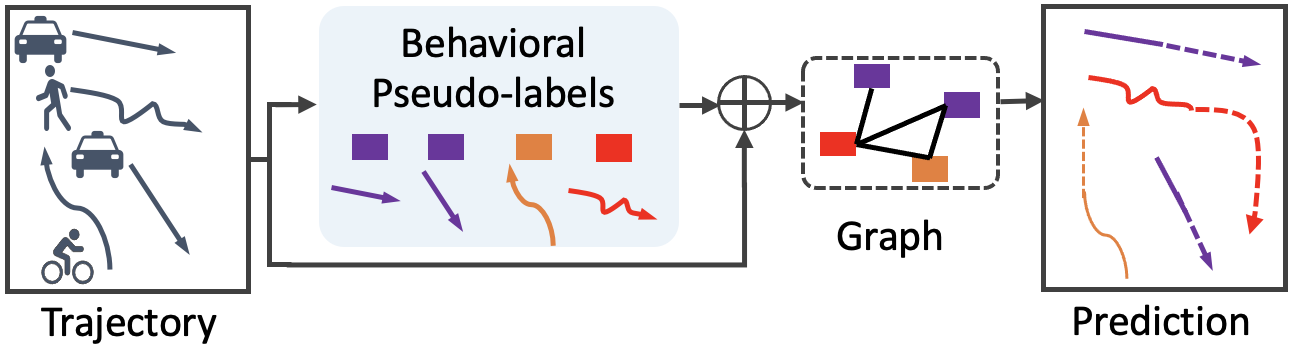}
  \caption{We propose the behavioral pseudo-labels learned from observed trajectories, effectively representing inter- and intra-type behavioral differences to improve pedestrian and heterogeneous trajectory prediction accuracy.}
  \label{fig:teaser}
\end{figure}

Existing trajectory prediction methods can be broadly classified into two categories. The first focuses on predicting \emph{pedestrian trajectories} in datasets that are exclusively composed of pedestrians \cite{pellegrini2009ETH, Lerner2007UCY} or deliberately omit non-pedestrian traffic agents \cite{mangalam2021ynet, zhao2021ExpTraj, mangalam2020PECNet}. These methods primarily employ neural networks to account for pedestrian social interactions, such as the pooling window mechanism \cite{Alexandre2016lstm} and social interaction graphs \cite{huang2019stgat, Mohamed2020socialstgcnn, shi2021sgcn}. The second category encompasses \emph{heterogeneous trajectory} prediction, considering a diverse range of traffic agents (e.g. cars, cyclists, pedestrians, etc.). Recent methods \cite{rainbow2021semanticStgcnn, fang2021UNIN, ruochen2022multiclassSGCN} exploit the annotated class labels of traffic agents to better model agent interactions in intricate urban scenarios. These labels facilitate the system's understanding on multifaceted interactions among various agent types \cite{ruochen2022multiclassSGCN}.

A notable research gap can be observed between pedestrian-only and heterogeneous trajectory prediction. Methods tailored solely for pedestrian behavior excel due to its predictable patterns but lack applicability in real-world scenarios like autonomous driving, since pedestrians behave very differently from heterogeneous agents \cite{rainbow2021semanticStgcnn, ruochen2022multiclassSGCN}. The fundamental differences in modeling the motion patterns of different types of agents stem from their distinct dynamics, speed ranges, spatial needs, interaction behaviors, decision-making processes, and ways of perceiving the environment, necessitating varied modeling approaches to accurately predict their trajectories. 
\RCCC{For heterogeneous trajectory prediction, ground-truth (GT) labels for agent types have traditionally been used to guide discriminative learning \cite{fang2021UNIN, ruochen2022multiclassSGCN, rainbow2021semanticStgcnn, du2024SFEM-GCN}. However, these labels often fail to capture diverse within-class behaviors: for example, ‘vans’ and ‘compact cars’ are both labeled simply as ‘cars,’ while ‘pedestrians’ can range from ordinary walkers to skateboarders \cite{Robicquet2016SDD}. This granularity issue can lead to mislabeling, especially when visually similar categories are grouped together. Moreover, obtaining such detailed GT labels is time-consuming and expensive. We argue that purely relying on manual labels is both insufficient and cost-ineffective for representing the nuanced motion patterns seen in real-world traffic scenarios.}

\RCC{In this paper, we present a unified framework utilizing machine-learned behavioral pseudo-labels applicable to both heterogeneous and  exclusively pedestrian domains. Our insight is that behavioral pseudo-labels can capture both inter-class and intra-class behavioral variations among agents, thereby improving the accuracy of our model. For heterogeneous scenarios, the use of behavioral pseudo-labels eliminates the need for manual label annotations, streamlining the process and reducing the reliance on extensive labeled datasets. In pedestrian-only scenarios, these pseudo-labels facilitate the differentiation and learning of intrinsic motion patterns among pedestrians, offering a more nuanced understanding of pedestrian behavior. A shared advantage across both contexts is the significant improvement in overall prediction performance, demonstrating the versatility and efficacy of behavioral pseudo-labels in diverse trajectory prediction tasks (\figurename~\ref{fig:teaser}).}

We propose the Behavioral Pseudo-Label informed Sparse Graph Convolution Network (BP-SGCN) for pedestrian and heterogeneous trajectory prediction. The network includes two modules. \RCCC{First, we introduce a deep unsupervised behavior clustering module that assigns pseudo-labels to agents based on their observed trajectories. This module marks a novel application of deep embedded clustering ~\cite{junyuan2015DEC}, utilizing high-level temporal latent features. It is supported by a Variational Recurrent Neural Network (VRNN) \cite{Junyoung2015VRNN} that processes a set of customized geometric features, crucial for capturing motion dynamics such as speed, angle, and acceleration. Additionally, a soft dynamic time warping loss addresses temporal variances in trajectories, uniquely tailoring our approach for trajectory modeling. The generated behavioral pseudo-labels are specifically designed to enhance trajectory forecasting, highlighting our model's focus on the nuanced demands of trajectory prediction in complex environments.}
\RCCC{Second, we propose a goal-guided pseudo-label informed trajectory prediction module, which adapts SGCN~\cite{shi2021sgcn}, a powerful GCN backbone for trajectory prediction that utilizes a sparse spatial-temporal attention mechanism to effectively model spatial interactions and temporal dependencies of agents. We then employ a Gumbel-Softmax straight-through estimator to link up the clustering module, allowing the prediction module and clustering module to be fine-tuned in an end-to-end manner.} \RCCC{Finally, we design a cascaded training scheme \cite{yang2023ManiCalibration} that first trains pseudo-label clustering in an unsupervised manner, and then fine-tunes both clustering and trajectory prediction together with the prediction loss to maximize their compatibility.}

BP-SGCN surpasses SOTAs in both heterogeneous prediction on the SDD \cite{Robicquet2016SDD} and Argoverse 1 \cite{mingfang2019argoverse} datasets, and in pedestrian prediction on the ETH/UCY \cite{pellegrini2009ETH, Lerner2007UCY} dataset and the pedestrian-only setup of SDD \cite{Becker2018AnEO}.
Our source code is available at \url{https://github.com/Carrotsniper/BP-SGCN} to facilitate further research. Our contributions are:
\begin{itemize}
  \item We propose the novel concept of behavioral pseudo-labels to represent clusters of traffic agents with different movement behaviors, improving trajectory prediction without the need for any extra annotation.

  \item To implement the idea, we propose BP-SGCN, which introduces a cascaded training scheme to optimize the compatibility of its two core modules: the pseudo-label clustering module and the trajectory prediction module. 
  \item We propose a deep unsupervised behavior clustering module to obtain behavioral pseudo-labels, tailoring the geometric feature representation and the loss to best learn the agents' behaviors.
  \item We propose a pseudo-label informed goal-guided trajectory prediction module, which facilitates end-to-end fine-tuning with its prediction loss for better clustering and prediction, outperforming existing pedestrian and heterogeneous prediction methods.
\end{itemize}

\section{Related work}

\subsection{Trajectory Prediction}
\label{sec:future}

Deep learning models have driven the latest advancement of trajectory prediction. 
Social-LSTM \cite{Alexandre2016lstm} introduces RNN-based neural networks \cite{HochSchm1997lstm} to model the trajectories of pedestrians and a pooling window mechanism to describe the interactions among them. Social-GAN~\cite{gupta2018socialgan} incorporates the ideas of Generative Adversarial Networks \cite{goodfellow2014GAN} to predict multiple multi-modal trajectories with distance-based interaction modeling. \textcolor{black}{Diffusion models \cite{chang2023design} are adopted into trajectory prediction \cite{gu2022stochastic, mao2023leapfrog}, showing significant improvement.}

Graph representations are increasingly recognized for their prowess in modeling relational features. TrafficPredict \cite{ma2019trafficPredict} incorporates soft-attention-based interaction graphs with LSTM to represent social interactions. STGAT \cite{huang2019stgat} models the trajectories using Spatial-Temporal Graph Attention Networks based on the sequence-to-sequence architecture. Social-STGCNN \cite{Mohamed2020socialstgcnn} introduces weighted graph edges, providing an interpretable measurement of pedestrian interactions. STAR \cite{cunjun2020star} takes advantage of the Transformer \cite{vaswani2017transformer} to construct a spatial-temporal graph transformer for trajectory representation. SGCN \cite{shi2021sgcn} advances this by proposing sparse directed spatial-temporal graph representations to model spatial interactions and motion tendencies for each pedestrian. However, these methods primarily focus on modeling pedestrian interactions, overlooking the intricate interactions among heterogeneous agents.

\RCCC{Future trajectory or goal information enhances the trajectory prediction as it provides valuable insights into the long-term intentions of individual agents~\cite{mangalam2020PECNet, mangalam2021ynet, zhao2021ExpTraj, ye2021agentformer, gu2022stochastic, Xu2022GroupNetMH, sun2021PCCSNet, pei2022socialVAE, xu2022memonet, li2024GoalOriented}. 
CVAE~\cite{sohn2015CVAE} based methods \cite{ye2021agentformer, pei2022socialVAE, Xu2022GroupNetMH} employ the future and past trajectory encoder during the training phase to train the latent representation for each agent.} Such latent is used to generate future trajectories during inference. Goal retrieval methods sample goal points and incorporate them as guides to model the prediction diversity~\cite{zhao2021ExpTraj, mangalam2021ynet, mangalam2020PECNet}. 
We employs the goal retrieval approach proposed in~\cite{zhao2021ExpTraj}, as it does not require training a separate models as in the CVAE-based method.

\RCCC{As trajectories are influenced by their surroundings, some studies employ image contextual features as auxiliary information. Convolutional Neural Networks (CNN) are applied to extract such features to improve prediction accuracy \cite{lee2017desire, wong2022V2net}. Semantic segmentation is adapted on scene images to extract annotated image features \cite{mangalam2021ynet}. While acknowledging the potential benefits of these techniques, we do not utilize scene-based features in this research. The integration of such features typically requires an auxiliary image processing network \cite{fang2023HSG, wong2022V2net, guo2022end2end-grid}, which complicates the model architecture and detracts from our primary focus on exploring the impact of behavioral pseudo-labels. Our approach, although not incorporating scene-based data, is conceptually aligned with these methods in its attempt to capture the nuanced behaviors and interactions of traffic agents purely from trajectory data.
The proposed method is orthogonal and complementary to the use of scene-based features. The two methods can be combined in future work.}

\begin{figure}[H]
  \centering
  \includegraphics[width=0.48\textwidth]{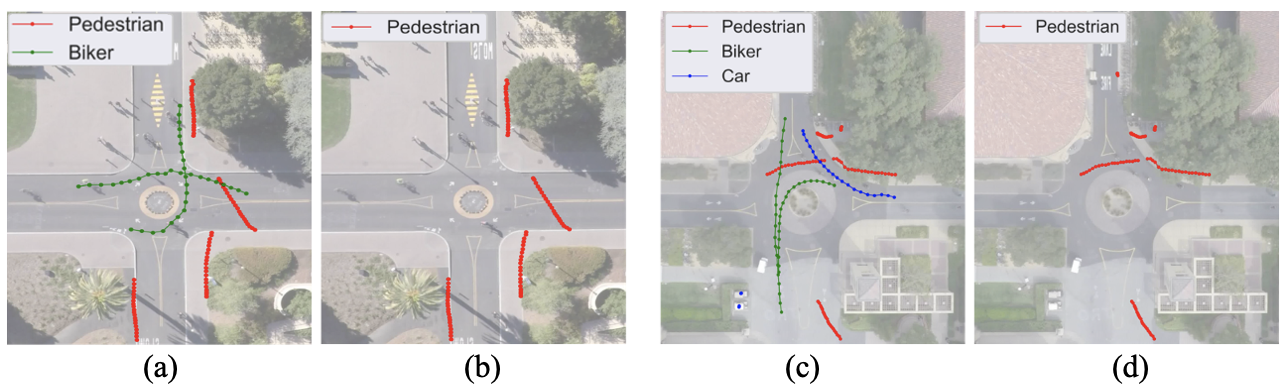}
  \caption{\RCC{Trajectory visualization on heterogeneous SSD dataset, where red, green and blue dots represent pedestrians, bikers and cars, respectively. (a) and (c) represent heterogeneous scenarios with all agent types, (b) and (d) represent the pedestrian-only scenarios commonly used by pedestrian trajectory predictions \cite{mangalam2020PECNet, mangalam2021ynet} by simply removing all non-pedestrian agents.}}
  \label{fig:pedestrian_traj}
\end{figure}
\subsection{Heterogeneous Trajectory Prediction}
Heterogeneous trajectory prediction considers traffic agents of all types. Adopting prediction approaches \cite{mangalam2021ynet, mohamed2022socialimplicit, Xu2022GroupNetMH} by simply ignoring non-pedestrian agents, or considering all agents to be of the same class, results in sub-optimal performance. Heterogeneous methods focus on modeling different agent behaviors. VP-LSTM~\cite{huikun2019JPKT} separately treats vehicles and pedestrians with LSTMs. Proposal-based approaches such as CoverNet \cite{Tung2020covernet} generate predefined multimodal trajectory anchors from observations of both vehicles and pedestrians. For better interpretability and representation of heterogeneous agent interactions, graph-based attention mechanisms \cite{vaswani2017transformer, huang2019stgat} are proposed. NLNI \cite{fang2021UNIN} presents a novel spatial-temporal category graph and proposed graph attention to capture the category-wise and agent-wise interactions. Multiclass-SGCN \cite{ruochen2022multiclassSGCN} and Semantic-STGCNN \cite{rainbow2021semanticStgcnn} introduce one-hot encoding to encode annotated class labels as part of node features. \RCCC{HIMRAE \cite{chen2024AEMTP} proposes dynamic interaction graphs among agents to reduce accumulated error for heterogeneous trajectory prediction. SMGCN \cite{zhang2024SMGCN} intorduces a sparse multi-relational GCN to learn heterogeneous interactions among agents.}

\RCCC{However, while these approaches achieve superior performance, most methods rely heavily on ground-truth labels to distinguish agent types. On the one hand, manual labeling is costly and error-prone, making the model’s performance overly dependent on label quality. On the other hand, focusing on semantic labels can overlook subtle behavioral differences among agents with the same semantic label. In this work, we propose learning behavioral pseudo-labels from agent motion dynamics, thereby reducing reliance on manual labels and capturing a broader spectrum of behaviors.}

\begin{figure*}[t]
  \includegraphics[width=\textwidth]{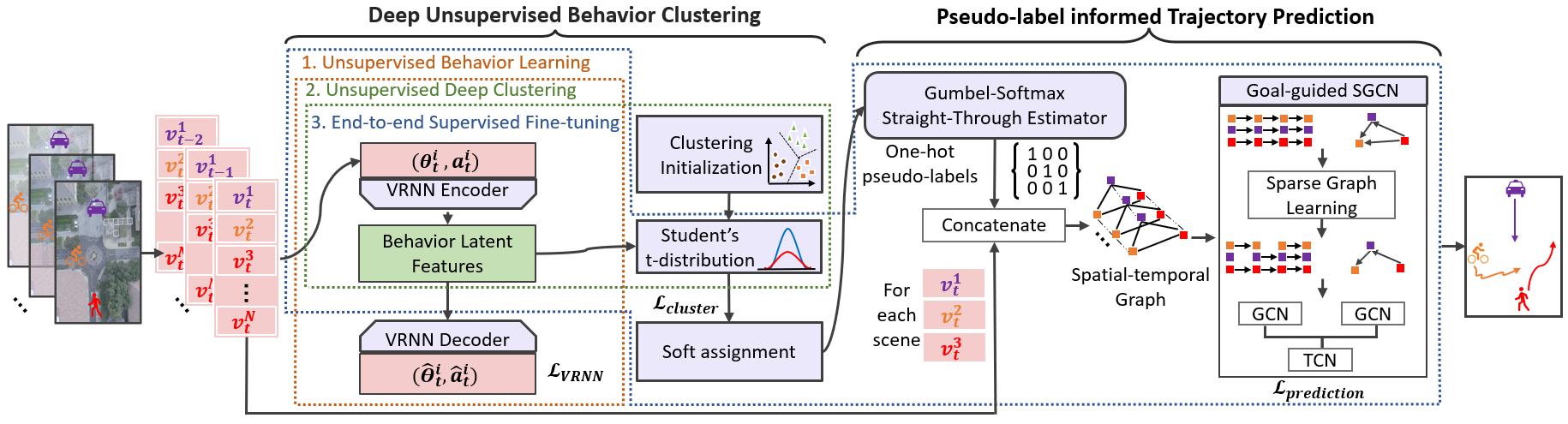}
  \caption{\RCCC{The overview of BP-SGCN to learn the pseudo-labels for trajectory prediction, consisting of  the deep unsupervised clustering module and the pseudo-label informed trajectory prediction module. We propose a cascaded optimization scheme to first learn pseudo-labels in an unsupervised manner, and then fine-tune them in an end-to-end manner with trajectory prediction supervision.}}
  \label{fig:overview}
\end{figure*}

\subsection{Motion Behavior Clustering}
\RCCC{The clustering of temporal trajectory patterns allows modeling the behavioral groups for better trajectory prediction \cite{fernando2018soft+, xue2020poppl}. Early works focus on the raw trajectory represented as 2D coordinates. 
Support vector clustering is introduced as a closed-loop method on motion vectors for motion behavior representations \cite{lawal2016support}. K-means on trajectory vectors or sequence key points obtain cluster centers to enhance trajectory prediction \cite{xue2020poppl, lui2021DDPTP}.}
DBSCAN is proposed to avoid manually specifying cluster numbers, adding more flexibility and interpretability to behavior patterns \cite{fernando2018soft+}. GP-Graph directly uses the absolute distance among pedestrians to determine the division of group \cite{bae2022gpgraph}. The recent PCCSNet leverages BiLSTM network to encode coordinates prior to K-means clustering, identifying behavioral modalities \cite{sun2021PCCSNet}. In addition to modalities, FEND further applies 1D CNN and LSTM for trajectory encoding and employs the K-means for long-tail trajectory clustering to distinguish trajectory patterns \cite{wang2023fend}.

\RCCC{However, most existing methods rely on shallow trajectory representations, limiting their ability to capture nuanced, evolving behaviors. Additionally, distance-based clustering approaches often struggle with complex motion patterns. To address these issues, we propose a cascaded optimization scheme featuring an end-to-end Deep Embedded Clustering (DEC) \cite{junyuan2015DEC} module, which iteratively refines cluster assignments using a KL-divergence objective. This dynamic adaptation yields richer latent representations, enabling a more data-driven and expressive approach to modeling agent behaviors.}

\section{Behavior Pseudo-Label Informed Sparse Graph Convolution Network}
\label{sec:methodology}

\subsection{The High-Level Network Architecture}

We observe a research gap in pedestrian and heterogeneous trajectory prediction. Existing pedestrian prediction approaches have limited applicability to heterogeneous traffic agents due to the diverse behaviors of agents. For instance, in \figurename~\ref{fig:pedestrian_traj}, (a) and (c) depict intricate heterogeneous scenarios with bikers and cars exhibiting longer, non-linear paths, while pedestrian-only scenarios (b) and (d) overlook interactions among pedestrians, bikers and cars.

Although introducing annotated class labels for heterogeneous agents leads to better prediction performance \cite{fang2021UNIN, ruochen2022multiclassSGCN, rainbow2021semanticStgcnn}, such labels are only a proxy of movement behaviors, which cannot represent intra-class behavioral differences and inter-class behavioral similarity.

To this end, we present the concept of \emph{behavioral pseudo-labels}, which capture movement behaviors to enhance trajectory prediction. Our pseudo-labels do not require annotations, mitigating the risk of mislabeling and reducing labor costs. It can be applied to both pedestrian-only and heterogeneous datasets, resulting in superior prediction performance. 

To realize pseudo-label informed trajectory prediction, we propose the \emph{Behavioral Pseudo-Label Informed Sparse Graph Convolution Network} (BP-SGCN). As shown in \figurename~\ref{fig:overview}, BP-SGCN includes two modules: deep unsupervised clustering and pseudo-label informed trajectory prediction. The former learns the pseudo-labels in an unsupervised manner, while the latter performs end-to-end optimization to improve pseudo-label clustering while predicting trajectories with such labels.

We propose the \emph{cascaded training scheme} to obtain the pseudo-labels and thus high-quality trajectory prediction. First, highlighted with the orange dotted block in \figurename~\ref{fig:overview}, the unsupervised behavior representation learning module derives behavior latent representations from observed trajectories through a Variational Recurrent Neural Network (VRNN) \cite{Junyoung2015VRNN} module. Then, in the green dotted block, the behavior latent representations are fed into simple clustering modules (e.g., K-means, GMM, etc.) for cluster center initialization. We then perform unsupervised deep clustering to learn the distribution of pseudo-labels by feeding the VRNN latent representations to the Student’s t-distribution kernel \cite{laurens2008tsne}. This allows fine-tuning the VRNN encoder to create a better latent space and refine the cluster centers. Finally, indicated by the blue dotted block, we utilize a Gumbel-Softmax straight-through estimator to sample one-hot pseudo-labels, which are concatenated to the trajectory features as the input of goal-guided SGCN \cite{shi2021sgcn} for trajectory prediction. The whole network is optimized end-to-end, fine-tuning the pseudo-label clustering module to maximize its compatibility for trajectory prediction.

\subsection{Deep Unsupervised Behavior Clustering}
Here, we explain how we obtain behavior clusters, which serve as powerful features for effective trajectory prediction.

\subsubsection{Geometric Representation of Trajectories}
Given a series of observed video frames of $N$ agents over time $t \in [ 1,T_{obs} ] $, and the corresponding 2-D trajectory coordinates $(x_{t}^{i}, y_{t}^{i})$, $ i \in [ 1,N ]$, our objective is to predict the future trajectory coordinates $p_{t}^{i} = (x_{t}^{i}, y_{t}^{i})$ of each traffic agent $i$ within a time horizon $t \in [T_{obs+1}, T_{pred} ]$. 

We introduce relative angle and acceleration magnitude to learn behavior latents. While global velocity is an effective feature for trajectory prediction~\cite{shi2021sgcn, ruochen2022multiclassSGCN}, it is less representative of behaviors, as it depends on global movement directions, and is less sensitive to velocity changes. Relative angles provide a representation that is invariant to the initial facing direction, which is complemented with the magnitude of acceleration that has been shown to be effective for modeling behaviors.

\RCC{For each traffic agent \textit{i}, we calculate its velocity vector at time $t$. For simplicity, we remove the notation \textit{i} in the following equation:
\begin{equation}
\vb*{v}_{t} = \left(\frac{x_{t} - x_{t-1}}{t - (t-1)}, \frac{y_{t} - y_{t-1}}{t - (t-1)}\right),
\label{velocity}
\end{equation}
where $\forall t\in[1,T_{obs}]$, we compute the cosine of the angle, $\cos{(\theta_{t})}$ between velocity vectors, $\vb*{v}_{t}$ and $\vb*{v}_{t-1}$:
\begin{equation}   
  \cos(\theta_{t}) = \frac { \vb*{v}_{t} \cdot  \vb*{v}_{t-1}}{| \vb*{v}_{t}| \cdot | \vb*{v}_{t-1}|}, \label{angle}
\end{equation}
and the magnitude of corresponding acceleration at time \textit{t}:
\begin{equation}
\left | \vb*{a}_{t} \right | = \left | \frac{\vb*{v}_{t} - \vb*{v}_{t-1}}{t-(t-1)}  \right | \label{acceleration}
\end{equation}    
The geometric feature is constructed as $g_{t} = (\cos(\theta_{t}), \left | \vb*{a}_{t} \right |)$.}

\subsubsection{Behavior Representation Learning}

We adapt VRNN to learn latent representations for behavior clustering \cite{sai2018DTC, junyuan2015DEC}. VRNN learns the temporal dependencies of a sequence by modeling the distribution over its hidden states with an encoder-decoder architecture. Compared to LSTM-based autoencoders \cite{sai2018DTC}, it effectively models the highly nonlinear dynamics and captures the uncertainties of latent space. Its probabilistic nature of variational inference improves the learning of implicit sequential data distributions.

In particular, the encoder network $\varphi_{enc}(\cdot,\cdot)$ receives the embedded geometric data $\varphi^{g}(g_{t})$ and recurrent hidden state $h_{t-1}$ to approximate the posterior distribution $q_{\phi}(\cdot)$: 
\begin{equation}
\begin{split}
    q_{\phi}(z_{t}|g_{\le t}, z_{< t}) = \mathcal{N} (z_{t}|(\mu _{z,t}, \sigma^{2}_{z,t})),\\
    [\mu _{z,t}, \sigma _{z,t}] = \varphi_{enc}(\varphi^{g}(g_{t}), h_{t-1}), 
\end{split}
\end{equation}
where $z_{t}$ is sampled using a reparameterization trick \cite{kingma2013VAE}. The decoder network $\varphi_{dec}(\cdot,\cdot)$ takes the embedded latent 
$\varphi^{z}(z_{t})$ and $h_{t-1}$ to approximate the reconstruction distribution $p_{\delta}(\cdot)$:
\begin{equation}
\begin{split}
    p_{\delta}(g_{t}|z_{\le t}, g_{< t}) = \mathcal{N} (g_{t}|(\mu _{g,t}, \sigma^{2}_{g,t})), \\
    [\mu _{g,t}, \sigma _{g,t}] = \varphi_{dec}(\varphi^{z}(z_{t}), h_{t-1}).
\end{split}
\end{equation}

To enhance the temporal dependencies in sequences, the prior distribution in VRNN relies on $h_{t-1}$ with $\varphi_{prior}(\cdot)$:
\begin{equation}
\begin{split}
    p_{\delta}(z_{t}|z_{< t}, g_{< t}) = \mathcal{N} (z_{t}|(\mu _{0,t}, \sigma^{2}_{0,t})), \\
    [\mu _{0,t}, \sigma _{0,t}] = \varphi_{prior}(h_{t-1}).
\end{split}
\end{equation}

We employ the Gated Recurrent Unit (GRU) \cite{Junyoung2014GRU} to update the RNN hidden state, which outperforms LSTM \cite{HochSchm1997lstm} when the sequence length is relatively short:
\begin{equation}
h_{t} = GRU(\varphi^{g}(g_{t}), \varphi^{z}(z_{t}), h_{t-1}).
\end{equation}
\RCC{
The VRNN is optimized with a customized loss:
\begin{equation}
   \mathcal{L}_{\textit{VRNN}}=  \mathcal{L}_{\textit{Soft-DTW}} + \mathcal{L}_{\textit{ELBO}},
\end{equation}
where $\mathcal{L}_{\textit{Soft-DTW}}$ is a differentiable soft Dynamic Time Warping (DTW) loss \cite{cuturi2017softdtw}:
\begin{equation}
   \mathcal{L}_{\textit{Soft-DTW}} = \underset{\mu _{g,t}}{min}\sum_{i=1}^{N}\frac{1}{T_{obs}}DTW_{\gamma}(\mu _{g,t}, g_{t}),
\end{equation}
$DTW_{\gamma}$ refers to the original DTW \cite{sakoe1978DTW} discrepancy that measures and aligns the similarity between two time series, $\gamma$ is a parameter indicating the acceptable distortion for aligning two sequences, $\mu_{g,t}$ is the decoded mean of the VRNN decoder.}
The loss allows capturing non-linear temporal alignment~\cite{zhao2018shapeDTW}, which cannot be achieved with MSE. 
$\mathcal{L}_{\text{ELBO}}$ is the variational evidence lower-bound with the Kullback–Leibler (KL) divergence \cite{kingma2013VAE, Junyoung2015VRNN}:
\begin{equation} \label{ELBO}
  \begin{split}
  \mathcal{L}_{\textit{ELBO}} = \mathbb{E}_{q_{\phi}(z_{\le T_{obs}} | g_{\le T_{obs}})} 
  \Biggl [  \sum_{t=1}^{T_{obs}} (\log_{}{p_{\delta}(g_{t} | z_{\le t}, g_{<t})} \\
  -KL(q_{\phi}( z_{t}| g_{\le t}, z_{<t})\ ||\ p_{\delta }(z_{t}|z_{< t}, g_{< t})) \Biggl ].
  \end{split}
\end{equation}
\RCCC{By optimizing $\mathcal{L}_{\textit{VRNN}}$, the model aligns predicted and observed sequences while maintaining a theoretically grounded variational framework. This alignment enhances flexibility in handling non-linear temporal dynamics, and the KL regularization constrains the latent structure, thus ensuring stable training. Consequently, the VRNN encoder provides richer latent representations for subsequent unsupervised deep clustering, effectively leveraging spatio-temporal structures to capture nuanced agent behaviors.}

\subsubsection{Deep Embedded Clustering}
We present a new application of Deep Embedded Clustering (DEC) \cite{junyuan2015DEC} to cluster the agent behaviors latents from the VRNN encoder, thereby generating a distribution of pseudo-labels. DEC allows jointly optimizing the cluster centers and the VRNN encoder, enhancing the latent representation via back-propagation. This significantly outperforms traditional methods like k-means \cite{MacQueen1967kmeans} and Gaussian mixture models \cite{Reynolds2009GMM}, which lack the capability to refine input feature representations.

The initial phase of DEC involves setting cluster centers using VRNN behavior latents. We input all training data into the VRNN encoder to obtain the set of behavior latent features $\mathbb{Z}$, and then apply k-means to determine initial centers, $c_{j}\in [1, k]$. Given the variance in agent behaviors across datasets, $k$ is an empirically tuned hyperparameter.

We then apply Student’s T-Distribution \cite{laurens2008tsne}, that is, Q distribution to compute the soft assignment between each initialized cluster center and latent vector \cite{junyuan2015DEC}. Its kernel measures the probability of each encoded vector $z_{i} \in \mathbb{Z}$ belonging to the cluster $j$: 
\begin{equation}
    q_{ij} = \frac{\left(1 + \frac{d(z_{i},c_j)}{\alpha}\right)^{-\frac{\alpha + 1}{2}}}{ \sum_{j^{\prime}}
    \left(1 + \frac{d(z_{i},c_{j^{\prime}})}{\alpha} \right)^{-\frac{\alpha +1}{2} }},
\end{equation}
where $d$ is a similarity metric that refers to the distance between the encoded vector $z_{i}$ and center $c_{j}$, and $\alpha$ is the number of degrees of freedom of the Q distribution. We denote $d$ as the Euclidean distance and set $\alpha$ to 1.

Meanwhile, we optimize the clustering network with a KL divergence loss to minimize the discrepancy between the two distributions:
\begin{equation}
    \mathcal{L}_{\text{cluster}} = KL(P||Q) =  {\textstyle \sum_{i}} {\textstyle \sum_{j}}\left(p_{ij}\log \frac{p_{ij}}{q_{ij}} \right), 
    \label{loss_cluster}
\end{equation}
where $P$ is the auxiliary distribution:
\begin{equation}
    p_{ij} = \frac{q_{ij}^{2}/f_{j}}{ {\textstyle \sum_{j^{\prime }}q_{ij^{\prime}}^{2}/f_{j^{\prime }}}},  \label{target distribution}
\end{equation}
and $f_{j} = \sum_{i}q_{ij}$ are soft cluster frequencies. \RCCC{Here, $P$ is re-weighted from Q distribution in a way that sharpens high-confidence assignments and de-emphasizes low-confidence ones \cite{junyuan2015DEC}, thereby systematically increasing the separation between clusters in the latent space. Finally, we derive soft assignments from the Student’s t-distribution, reflecting the probability of the latent $z^{i}_t$ in each cluster $c_j$. This approach not only offers greater flexibility in representing complex behaviors but also sharpens cluster boundaries by reinforcing high-confidence assignments and reducing ambiguity in low-confidence ones. Consequently, it yields more coherent clusters and better captures the inherent diversity in agent dynamics, ultimately enhancing the overall clustering quality.}

\subsection{Pseudo-label Informed Trajectory Prediction}
Here, we introduce the concept of behavioral pseudo-labels for more accurate trajectory prediction. 

\subsubsection{Gumbel-Softmax Straight-Through Estimator}
While the soft assignment represents good behavior clusters, such clusters are unsupervised and trained only on feature representations, meaning that they are still sub-optimal for any given task. This explains the sub-optimal prediction accuracy in existing methods \cite{sun2021PCCSNet, wang2023fend}. Here, we present a framework to improve the compatibility between the clusters and the task via fine-tuning the behavior latent.  

To enable end-to-end fine-tuning of the behavior latent with a task objective, an operator is needed to connect the clustering and the prediction modules. We employ the Gumbel-Softmax straight-through estimator \cite{Jang2016gumbelsoftmax}, which facilitates the gradient propagation and computes one-hot vectors representing the pseudo-labels. The estimator uses a differentiable Softmax, as opposed to the non-differentiable Argmax, allowing end-to-end optimization. An agent's class label is $l_{j}$, where $j\in 1,\dots k$ is the cluster center.

Apart from performance gains, as one-hot labels fit the human understanding of a class concept, they allow better interpretability via visualization tools. They are also immediately compatible with existing network architectures trained with ground-truth labels \cite{rainbow2021semanticStgcnn,ruochen2022multiclassSGCN}, allowing effective adaptations.

\subsubsection{Behavioral Pseudo-Label Informed SGCN}
\RCCC{We adopt a Sparse Graph Convolution Network (SGCN) \cite{shi2021sgcn} as our backbone and introduce the pseudo-labels and a new loss function. SGCN has shown outstanding performance and is computationally efficient on pedestrian trajectory prediction~\cite{shi2021sgcn}. It introduces sparsified spatial-temporal attention mechanism \cite{vaswani2017transformer, li2021temporalpyramid, huang2019stgat}, which effectively models spatial interactions and temporal dependencies among agents.} The sparse graph learning component removes spatial superfluous interactions and temporal motion tendencies, improving both computational speed and accuracy. In reality, our pseudo-label framework is compatible with a wide range of trajectory prediction networks.

We introduce the usage of semantic-goal features into SGCN, which enhances the prediction accuracy \cite{mangalam2021ynet, zhao2021ExpTraj}. To this end, we integrate the goal-retrieval operation \cite{zhao2021ExpTraj} into the SGCN, we first subtract each observation step $\vb*{v}_{t}$ in $t \in [1, T_{obs}]$ by the corresponding trajectory endpoint $\vb*{v}_{T_{pred}}$ as $\vb*{v}_{t} = \vb*{v}_{t} - \vb*{v}_{T_{pred}}$. \RCC{We then construct the spatial graph $\mathcal{G}_{s} = \left \{  (\mathcal{V}_{s}, \mathcal{A}_{s})  | \mathcal{V}_{s} \in \mathbb{R}^{T_{obs} \times N \times D_{s}}, \mathcal{A}_{s} \in \mathbb{R}^{T_{obs} \times N \times N}\right \}$, where $\mathcal{V}_{s}$ represents the spatial interactions among all agents at time step $t$, $\mathcal{A}_{s}$ is the spatial adjacency matrix and $D_{s}$ refers to the spatial feature dimension.}

\textcolor{black}{To add heterogeneity to the graph, we concatenate the pseudo-labels $l$ to the trajectory feature vector for each agent at each time step as $\mathcal{V}^{i}_{t} = \textit{concat}({v}^{i}_{t}, l^{i})$, $\forall t \in [1, T_{obs}]$ and $\forall i \in [1, N]$. \RCC{Similarly, we establish the temporal graph $\mathcal{G}_{t} = \left \{(\mathcal{V}_{t}, \mathcal{A}_{t}) | \mathcal{V}_{t} \in \mathbb{R}^{N \times T_{obs} \times D_{t}}, \mathcal{A}_{t} \in \mathbb{R}^{N \times T_{obs} \times T_{obs}} \right \}$ to represent the temporal correlations of each individual agent during $T_{obs}$ steps, where $\mathcal{A}_{t}$ is the temporal adjacency matrix and $D_{t}$ is the temporal feature dimension}. Finally, these spatial and temporal goal-guided heterogeneous graphs are passed into SGCN for final trajectory prediction.}

We propose a joint training strategy with a novel loss function to jointly optimize trajectory prediction and pseudo-label clustering. Thanks to our Gumbel-Softmax estimator, back-propagation is performed from the prediction all the way back to the VRNN encoder, resulting in better compatibility between the clustering and prediction modules. We present a combined loss:
\begin{equation}
    \mathcal{L}_{final} = \mathcal{L}_{cluster} + \mathcal{L}_{prediction}, 
\end{equation}
where $\mathcal{L}_{cluster}$ is defined in Eq. \ref{loss_cluster}, and $\mathcal{L}_{prediction}$ as:
\begin{equation}
    \mathcal{L}_{prediction} = - {\textstyle \sum_{t = T_{obs}+1}^{T_{pred}}}\log P(p_{t}|\hat{\mu}, \hat{\sigma },\hat{\rho  }),
\end{equation}
where $\hat{\mu}$ and $\hat{\sigma}$ are the mean and variance of the bi-variate Gaussian distribution of trajectory prediction, and $\hat{\rho}$ represents the correlation coefficient.

\section{Experiments}

\subsection{Datasets}

We evaluate BP-SGCN on multiple benchmark datasets, including the Stanford Drone Dataset (SDD) \cite{Robicquet2016SDD}, Argoverse 1~\cite{mingfang2019argoverse}, ETH \cite{pellegrini2009ETH} and UCY \cite{Lerner2007UCY}, and the pedestrian-only version of SDD \cite{Becker2018AnEO}. For pedestrian trajectory prediction, ETH/UCY consists of five pedestrian-only datasets (ETH, HOTEL, UNIV, ZARA1, ZARA2) with 1,536 pedestrians. Pedestrian-only SDD is the simplified version where non-pedestrian agents are removed. For heterogeneous trajectory prediction, we follow \cite{guo2022end2end-grid, junwei2020multiverse, ruochen2022multiclassSGCN} that consider all trajectories, consisting of 8 scenes, 60 videos and 6 categories of traffic agents (i.e., pedestrians, bicyclists, skateboarders, carts, cars, and buses). Argoverse 1 consists of over 30K urban traffic scenarios that include 3 types of agents (i.e. AVs, agents, and others).

\subsection{Experimental Setup}
By default, we follow the experimental setup of \cite{shi2021sgcn, mohamed2022socialimplicit}, using 3.2 seconds (8 frames) of observation trajectories to predict the next 4.8 seconds (12 frames). For pedestrian-only prediction, we employ the data augmentation approach introduced in \cite{mohamed2022socialimplicit} and the official leave-one-out strategy \cite{gupta2018socialgan} during the training and validation.
\RCCC{For heterogeneous trajectory prediction on Argoverse 1 dataset, our experimental setup and dataset split strategy follow \cite{fang2021UNIN, fang2023HSG}. Specifically, we utilize 2 seconds (20 frames) of observation trajectories to predict the trajectories of all tracked objects over the subsequent 3 seconds (30 frames) within each scene. In particular, our experimental setup on the Argoverse 1 dataset for heterogeneous trajectory prediction predicts trajectories for all agents \cite{fang2021UNIN, zhang2023bip-tree}, unlike methods focusing on a single agent \cite{zhou2022Hivt} or two specific agents \cite{li2024GoalOriented}, our approach captures multi-agent interactions, reflecting real-world traffic complexity and improving predictive robustness, situational awareness, and adaptability to diverse urban environments.}

\RCCC{During testing, we adhere to the standard protocol by generating 20 predictions for both heterogeneous \cite{fang2023HSG, rainbow2021semanticStgcnn, ruochen2022multiclassSGCN} and pedestrian-only trajectory predictions \cite{gupta2018socialgan, monti2021dagnet, shi2021sgcn}. This approach ensures our results are comparable to those established in the field. The sample with the lowest error is then used to compute the evaluation metrics.}
We employ the Average Displacement Error (ADE) and Final Displacement Error (FDE) \cite{Alexandre2016lstm, gupta2018socialgan, Mohamed2020socialstgcnn, shi2021sgcn} as our evaluation metrics: 
\begin{equation}
\label{metrics}
\begin{split}
    &ADE = \frac{1}{(T_{pred}-T_{obs}) \times N}  \sum_{i=1}^{N} {\sum_{t=T_{obs}+1}^{T_{pred}}||\hat{p}_{t}^{i} - p_{t}^{i}||_{2}} ,  \\
    &FDE = \frac{1}{N} {\sum_{i=1}^{N} ||\hat{p}_{t}^{i} - p_{t}^{i}||_{2}}, t = T_{pred},
\end{split}
\end{equation}
where $\hat{p}_{t}^{i}$ represents the ground-truth trajectory coordinates.

\begin{table}[!t]
\scriptsize
  \begin{center}
  \renewcommand{\arraystretch}{1.10}
  \setlength{\tabcolsep}{2.5pt}
  \caption{Results on SDD for heterogeneous prediction.}
  \begin{tabular}{*{2}{c}cc|cc}
    \hline
    \label{tab:SDD_results}
    \multirow{2}{*}{Methods}   &\multirow{2}{*}{Venue} &\multirow{2}{*}{Year} &\multirow{2}{*}{GT Labels} & \multicolumn{2}{c}{ SDD } \\
    &  &  & &ADE($\downarrow$) &FDE($\downarrow$) \\
    \hline\hline
    Social-LSTM \cite{Alexandre2016lstm}                  &CVPR      &2016 &No    &31.19  &56.97  \\
    DESIRE \cite{lee2017desire}                          &CVPR      &2017 &No    &19.25  &34.05  \\
    MATF \cite{tianyang2019MATF}                         &CVPR      &2019 &No    &22.59  &33.53  \\
    STGAT \cite{huang2019stgat}                           &ICCV      &2019 &No    &18.80  &31.30  \\
    Multiverse \cite{junwei2020multiverse}               &CVPR      &2020 &No    &14.78  &27.09  \\
    SimAug \cite{liang2020simaug}                        &ECCV      &2020 &No    &10.27  &19.71  \\
    NLNI \cite{fang2021UNIN}                               &ICCV      &2021 &Yes   &15.90  &26.30  \\
    STSF-Net \cite{wang2023STSF-Net}                      &TMM       &2021 &No    &14.81  &28.03
    \\
    Semantic-STGCNN  \cite{rainbow2021semanticStgcnn}     &SMC       &2021 &Yes   &18.12  &29.70  \\
    $V^{2}$-Net \cite{wong2022V2net}                     &ECCV    &2022 &No    &\underline{7.12}   &\underline{11.39}  \\
    Multiclass-SGCN \cite{ruochen2022multiclassSGCN}      &ICIP      &2022 &Yes   &14.36  &25.99  \\
    TDOR \cite{guo2022end2end-grid}                      &CVPR      &2022 &No    &8.60   &13.90  \\

    \RCC{CAPHA} \cite{azadani2023capha}               &\RCC{TVT}   &\RCC{2023} &\RCC{No}    & \RCC{9.13}   &\RCC{14.34}  \\

    \RCCC{VNAGT} \cite{chen2023NVAGT}               &\RCCC{TVT}   &\RCCC{2023} &\RCCC{Yes}    & \RCCC{9.67}   &\RCCC{17.22}  \\

      \RCCC{SFEM-GCN} \cite{du2024SFEM-GCN}               &\RCCC{TIV}   &\RCCC{2024} &\RCCC{Yes}    & \RCCC{15.31}   &\RCCC{25.72}  \\

    \RCCC{SMGCN} \cite{zhang2024SMGCN}               &\RCCC{IJCAI}   &\RCCC{2024} &\RCCC{Yes}    & \RCCC{20.89}   &\RCCC{36.84}  \\

    \hline
    BP-SGCN (Ours)                                        &          &     &No    &\textbf{6.94} &\textbf{9.57} \\
    \hline
  \end{tabular}

  \end{center}
\end{table}

\begin{table}[!t]
\scriptsize
  \begin{center}
  \renewcommand{\arraystretch}{1.10}
  \setlength{\tabcolsep}{2.5pt}
  \caption{Results on Argoverse 1 for heterogeneous prediction. 
  }
  \begin{tabular}{*{2}{c}cc|cc}
    \hline
    \label{tab:argoverse}
    \multirow{2}{*}{Methods}   &\multirow{2}{*}{Venue} &\multirow{2}{*}{Year} &\multirow{2}{*}{GT Labels} & \multicolumn{2}{c}{ Argoverse 1 } \\
    &  &  & &ADE($\downarrow$) &FDE($\downarrow$) \\
    \hline\hline
    Social-LSTM \cite{Alexandre2016lstm}                 &CVPR      &2016 &No    &1.39      &2.57 \\
    DESIRE \cite{lee2017desire}                         &CVPR      &2017 &No    &0.90      &1.45 \\
    R2P2-MA\cite{Rhinehart2018r2p2AR}                   &ECCV      &2018 &No    &1.11      &1.77 \\
    MATFG \cite{tianyang2019MATF}                      &CVPR      &2019 &No    &1.26      &2.31 \\ 
    CAM\cite{Park2020DiverseAA}                         &ECCV      &2020 &No    &1.13      &2.50 \\
    MFP\cite{Tang2019MultipleFP}                        &NeurIPs   &2020 &No    &1.40      &2.68 \\
    Social-STGCNN \cite{Mohamed2020socialstgcnn}         &CVPR      &2020 &No    &1.31      &2.34 \\
    NLNI \cite{fang2021UNIN}                             &ICCV      &2021 &Yes   &0.79      &1.26 \\

    DD \cite{xu2022DD}                                  &Inf. Sci. &2022 &No &\underline{0.74}
    &1.28 \\
    
    HRG+HSG \cite{fang2023HSG}                          &TITS      &2023 &No    &0.85      &\textbf{1.12} \\ 
    BIP-Tree \cite{zhang2023bip-tree}                    &TITS      &2023 &No    &0.78      &1.35 \\
    \hline
    BP-SGCN (Ours)  &          &     &No    &\textbf{0.69}   &\underline{1.15} \\
    \hline
  \end{tabular}

  \end{center}
\end{table}

\begin{table*}[t]
\scriptsize
  \centering
  \begin{threeparttable}[b]
  \renewcommand{\arraystretch}{1.12}
  \caption{\textcolor{black}{Results on ETH/UCY on pedestrian-only prediction; \\- denotes missing result due to unavailability from original authors.}}
  \label{tab:ethanducy}
  \begin{tabular}{*{2}{c}c|cccccc}
    \hline
    \multirow{2}{*}{Method} &\multirow{2}{*}{Venue} &\multirow{2}{*}{Year} 
    &\multicolumn{1}{c}{ETH} & \multicolumn{1}{c}{HOTEL} & \multicolumn{1}{c}{UNIV} & \multicolumn{1}{c}{ZARA1} & \multicolumn{1}{c}{ZARA2} & \multicolumn{1}{c}{AVG} \\
    &  &  & \RCC{
    ADE($\downarrow$)/FDE($\downarrow$)} &\RCC{ADE($\downarrow$)/FDE($\downarrow$)} &\RCC{ADE($\downarrow$)/FDE($\downarrow$)} &\RCC{ADE($\downarrow$)/FDE($\downarrow$)} &\RCC{ADE($\downarrow$)/FDE($\downarrow$)}  &\RCC{ADE($\downarrow$)/FDE($\downarrow$)} \\
    \hline\hline
    Social LSTM \cite{Alexandre2016lstm}               &CVPR    &2016  &1.09/2.35 &0.79/1.76 &0.67/1.40 &0.47/1.00 &0.56/1.17 &0.72/1.54 \\
    Social GAN \cite{gupta2018socialgan}               &CVPR    &2018  &0.87/1.62 &0.67/1.37 &0.76/1.52 &0.35/0.68 &0.42/0.84 &0.61/1.21 \\
    Social-STGCNN \cite{Mohamed2020socialstgcnn}       &CVPR    &2020  &0.64/1.11 &0.49/0.85 &0.44/0.79 &0.34/0.53 &0.30/0.48 &0.44/0.75 \\
    PECNet \cite{mangalam2020PECNet}                   &ECCV    &2020  &0.54/0.87 &0.18/0.24 &0.35/0.60 &0.22/0.39 &0.17/0.30 &0.29/0.48      \\
    SGCN \cite{shi2021sgcn}                            &CVPR    &2021  &0.63/1.03 &0.32/0.55 &0.37/0.70 &0.29/0.53 &0.25/0.45 &0.37/0.65 \\
    AgentFormer \cite{ye2021agentformer}               &ICCV    &2021  &0.45/0.75  &0.14/0.22 &0.25/0.45 &0.18/0.30 &0.14/0.24 &0.23/0.39      \\
    PCCSNet \cite{sun2021PCCSNet}                      &ICCV    &2021  &\underline{0.28}/0.54  &0.11/0.19 &0.29/0.60 &0.21/0.44 &0.15/0.34 &0.21/0.42      \\
    \RCC{ExpertTraj \tnote{1}} \cite{zhao2021ExpTraj}                  &\RCC{ICCV}    &\RCC{2021}  &\RCC{0.37/0.65} &\RCC{0.11/\underline{0.15}} &\RCC{0.20/0.44} &\RCC{0.15/0.31} &\RCC{0.12/0.25} &\RCC{0.19/0.36}      \\
    
    STSF-Net \cite{wang2023STSF-Net}                   &TMM     &2021  &0.63/1.13 &0.24/0.43 &0.28/0.52 &0.23/0.45 &0.21/0.41 &0.32/0.59 \\

    Social-Implicit \cite{mohamed2022socialimplicit}   &ECCV    &2022  &0.66/1.44 &0.20/0.36 &0.31/0.60 &0.25/0.50 &0.22/0.43 &0.33/0.67 \\
    GP-Graph \cite{bae2022gpgraph}                     &ECCV    &2022
    &0.43/0.63 &0.18/0.30 &0.24/0.42 &0.17/0.31 &0.15/0.29 &0.23/0.39 \\
    
    Social-VAE \cite{pei2022socialVAE}                 &ECCV    &2022  &0.41/0.58  &0.13/0.19 &0.21/0.36 &0.17/0.29 &0.13/0.22 &0.21/0.33      \\

    MemoNet \cite{xu2022memonet}                       &CVPR    &2022  &0.40/0.61  &0.11/0.17 &0.24/0.43 &0.18/0.32 &0.14/0.24 &0.21/0.35      \\
    
    GroupNet \cite{Xu2022GroupNetMH}                   &CVPR    &2022  &0.46/0.73 &0.15/0.25 &0.26/0.49 &0.21/0.39 &0.17/0.33 &0.25/0.44      \\

    MID \cite{gu2022stochastic}                        &CVPR    &2022  &0.39/0.66 &0.13/0.22 &0.22/0.45 &0.17/0.30 &0.13/0.27 &0.21/0.38      \\

    \RCCC{GTPPO} \RCCC{\cite{yang2022GTPPO}}                     &\RCCC{TNNLS}  &\RCCC{2022}
    &\RCCC{0.63/0.98} &\RCCC{0.19/0.30} &\RCCC{0.35/0.60} &\RCCC{0.20/0.32} &\RCCC{0.18/0.31} &\RCCC{0.31/0.50} \\

    Graph-TERN \cite{bae2023TERN}                      &AAAI    &2023  &0.42/0.58 &0.14/0.23 &0.26/0.45 &0.21/0.37 &0.17/0.29 &0.24/0.88      \\

    MSRL \cite{wu2023MSRL}                             &AAAI    &2023  &\underline{0.28}/\underline{0.47} &0.14/0.22 &0.24/0.43 &0.17/0.30 &0.14/0.23 &0.19/0.33      \\ 

    LED \cite{mao2023leapfrog}                         &CVPR    &2023  &0.39/0.58 &0.11/0.17 &0.26/0.43
    &0.18/0.26 &0.13/0.22 &0.21/0.33      \\

    EqMotion \cite{xu2023eqmotion}                     &CVPR    &2023  &0.40/0.61 &0.12/0.18 &0.23/0.43 &0.18/0.32 &0.13/0.23 &0.21/0.35       \\

    FEND \cite{wang2023fend}                           &CVPR    &2023  &-         &-         &-         &-         &-          &\underline{0.17}/0.32      \\

    EigenTrajectory \cite{bae2023eigentrajectory}      &ICCV    &2023  &0.36/0.53 &0.12/0.19 &0.24/0.43 &0.19/0.33 &0.14/0.24 &0.21/0.34      \\
    
    \RCC{TUTR} \cite{shi2023TUTR}                      &\RCC{ICCV}    &\RCC{2023}
    &\RCC{0.40/0.61} &\RCC{0.11/0.18} &\RCC{0.23/0.42} &\RCC{0.18/0.34} &\RCC{0.13/0.25} &\RCC{0.21/0.36} \\

    \RCC{SICNet} \cite{dong2023sparse}                 &\RCC{ICCV}    &\RCC{2023}
    &\RCC{\textbf{0.27}/\textbf{0.45}} &\RCC{0.11/0.16} &\RCC{0.26/0.46} &\RCC{0.19/0.33} &\RCC{0.13/0.26} &\RCC{0.19/0.33} \\

    \RCC{TP-EGT} \cite{yang2024TP-EGT}                 &\RCC{TITS}    &\RCC{2023}
    &\RCC{0.41/0.68} &\RCC{0.13/0.21} &\RCC{0.29/0.50} &\RCC{0.18/0.30} &\RCC{0.16/0.27} &\RCC{0.23/0.39} \\

    \RCC{DynGroupNet} \cite{xu2024DYNAMICGROUP}      &\RCC{NN}    &\RCC{2023}
    &\RCC{0.42/0.66} &\RCC{0.13/0.20} &\RCC{0.24/0.44} &\RCC{0.19/0.34} &\RCC{0.15/0.28} &\RCC{ 0.23/0.38} \\

    \RCC{SMEMO} \cite{marchetti2024smemo}      &\RCC{TPAMI}    &\RCC{2024}
    &\RCC{0.39/0.59} &\RCC{0.14/0.20} &\RCC{0.23/0.41} &\RCC{0.19/0.32} &\RCC{0.15/0.26} &\RCC{0.22/0.35} \\

     \RCCC{STGlow} \RCCC{\cite{liang2023stglow}}                      &\RCCC{TNNLS}   &\RCCC{2024}  &\RCCC{0.31/0.49} &\RCCC{\textbf{0.09}/\textbf{0.14}} &\RCCC{\textbf{0.16}/\underline{0.33}} &\RCCC{\textbf{0.12}/\underline{0.24}} &\RCCC{\textbf{0.09}/\underline{0.19}} &\RCCC{\textbf{0.15}/\underline{0.28}} \\

    \RCCC{MRGTraj} \cite{peng2024mrgtraj} &\RCCC{TCSVT}    &\RCCC{2024}  
    &\RCCC{\underline{0.28}/\underline{0.47}} &\RCCC{0.21/0.39} &\RCCC{0.33/0.60} &\RCCC{0.24/0.44} &\RCCC{0.22/0.41} 
    &\RCCC{0.26/0.46} \\

    \RCCC{HighGraph} \cite{kim2024higher} &\RCCC{CVPR} &\RCCC{2024} 
    &\RCCC{0.40/0.55} &\RCCC{0.13/0.17} &\RCCC{0.20/\underline{0.33}} &\RCCC{0.17/0.27} &\RCCC{0.11/0.21} &\RCCC{0.20/0.30} \\

    \RCCC{PPT} \cite{lin2024progressivePPT} &\RCCC{ECCV}         &\RCCC{2024}
    &\RCCC{0.36/0.51} &\RCCC{0.11/\underline{0.15}} &\RCCC{0.22/0.40} &\RCCC{0.17/0.30} &\RCCC{0.12/0.21} &\RCCC{0.20/0.31} \\

    \hline   
    BP-SGCN (Ours)                              &       &     &0.33/\underline{0.47} &\underline{0.10}/\textbf{0.14} &\underline{0.17}/\textbf{0.26} &\underline{0.13}/\textbf{0.19} &\underline{0.10}/\textbf{0.16} &\underline{0.17}/\textbf{0.24}\\       
    \hline
    \multicolumn{1}{c}{} & \\ 
  \end{tabular}
     \begin{tablenotes}
     \item[1] \RCC{For ExpertTraj\cite{zhao2021ExpTraj}, the discrepancy from the original paper arises due to an error highlighted by the authors: \url{https://github.com/JoeHEZHAO/expert_traj}}
   \end{tablenotes}
  \end{threeparttable}
\end{table*}


\begin{table}[!h]
\scriptsize
  \begin{center}
  \renewcommand{\arraystretch}{1.10}
  \setlength{\tabcolsep}{2.5pt}
  \caption{Results on the pedestrian-only version of SDD.}
  \begin{tabular}{*{2}{c}c|cc}
    \hline
    \label{tab:sddnolabel}
    \multirow{2}{*}{Methods}   &\multirow{2}{*}{Venue} &\multirow{2}{*}{Year} & \multicolumn{2}{c}{ SDD-human } \\
    &   & &ADE($\downarrow$) &FDE($\downarrow$) \\
    \hline\hline
    STGAT \cite{huang2019stgat}                       &ICCV      &2019  &0.58  &1.11 \\
    Social-Ways \cite{amirian2019socialways}          &CVPRW     &2019  &0.62  &1.16 \\
    DAG-Net \cite{monti2021dagnet}                    &ICPR      &2020  &0.53  &1.04 \\
    Social-implicit \cite{mohamed2022socialimplicit}  &ECCV      &2022  &0.47  &0.89 \\
    \RCCC{WTGCN \cite{chen2024wtgcn}}                 &\RCCC{IJMLC}   &\RCCC{2024}  &\RCCC{\underline{0.43}}  &\RCCC{0.72} \\
    \RCCC{IGGCN \cite{CHEN2025104862}}                 &\RCCC{DSP}   &\RCCC{2024}  &\RCCC{0.44}  &\RCCC{\underline{0.71}} \\
    
    \hline
    BP-SGCN (Ours)            &     &    &\textbf{0.28}         &\textbf{0.41}    \\
    \hline
  \end{tabular}
  \end{center}
\end{table}

\subsection{Quantitative Evaluation}

\subsubsection{Heterogeneous Prediction}
\RCCC{\tablename~\ref{tab:SDD_results} compares our BP-SGCN with previous state-of-the-art methods on heterogeneous SDD. These methods can be categorized into three groups based on the input features, including trajectory-only \cite{Alexandre2016lstm, huang2019stgat, wang2023STSF-Net, bae2022gpgraph, du2024SFEM-GCN}, trajectory with ground-truth labels \cite{fang2021UNIN, rainbow2021semanticStgcnn, ruochen2022multiclassSGCN, chen2023NVAGT, du2024SFEM-GCN, zhang2024SMGCN}, and trajectory with extra scene features such as scene semantics \cite{lee2017desire, junwei2020multiverse, liang2020simaug, wong2022V2net, guo2022end2end-grid, azadani2023capha}. BP-SGCN outperforms all the methods that utilize ground-truth agent class labels \cite{ruochen2022multiclassSGCN,fang2021UNIN,rainbow2021semanticStgcnn,chen2023NVAGT, du2024SFEM-GCN, zhang2024SMGCN}. Compared to the best method VNAGT \cite{chen2023NVAGT}, BP-SGCN demonstrates the superiority by reducing ADE/FDE by 28.23\%/44.43\%. Crucially, for SOTA approaches that incorporate scene semantic features such as $V^{2}$-Net \cite{wong2022V2net} and TDOR \cite{guo2022end2end-grid}, our BP-SGCN improves the performance by reducing ADE/FDE by 2.5\%/15.9\% compared to $V^{2}$-Net and 19.3\%/31.2\% compared to TDOR. The results indicate that without the need for additional inputs, our BP-SGCN can still achieve SOTA performance in heterogeneous trajectory prediction.}

\tablename~\ref{tab:argoverse} compares the BP-SGCN with those state-of-the-art methods in heterogeneous trajectory prediction on Argoverse 1, following the setup in \cite{fang2021UNIN, fang2023HSG, zhang2023bip-tree}. Results show that our BP-SGCN outperforms all the methods by a significant margin, especially in the ADE metric. \RCCC{BP-SGCN surpasses NLNI \cite{fang2021UNIN}, which integrates ground-truth labels, by reducing 12.7\% in ADE and 8.7\% in FDE, further showcasing the effectiveness of our proposed pseudo-label module. Notably, although NLNI utilizes label-based category features, its performance is limited by the simplistic nature of the ``GT Labels" in the Argoverse 1 dataset, which are broadly classified as ``1 AV" (1 Autonomous Vehicle), ``1 Focal" (the primary vehicle whose trajectory is predicted), and ``N other" (other tracked objects, which can include vehicles, pedestrians, or bicycles). This coarse categorization restricts the algorithm's ability to accurately capture and analyze the nuanced interactions among diverse traffic agents. In contrast, BP-SGCN effectively overcomes these constraints by conducting a comprehensive analysis of the behavior dynamics of all agents within the scene. By employing our advanced pseudo-label module, we significantly enhance the representational capabilities of our system, leading to markedly improved prediction accuracy across diverse traffic scenarios. This improvement is achieved without the need for direct matching with ground-truth labels, demonstrating the robustness and adaptability of our approach in interpreting complex interactive behaviors.} \RCCC{Importantly, DD \cite{xu2022DD} and HRG+HSG\cite{fang2023HSG} achieve comparable performance on ADE and FDE mainly due to the use of scene images that better capture the interactions between traffic agents and environments, our BP-SGCN still shows the best ADE performance compared to these methods.}

\subsubsection{Pedestrian Prediction}

For ETH/UCY, we conduct quantitative comparisons with a wide range of methods with various techniques, as shown in \tablename~\ref{tab:ethanducy}. Following \cite{xu2023eqmotion, mao2023leapfrog}, we compare with methods utilizing trajectory data only.

\RCC{For distribution-based methods, Social-LSTM \cite{Alexandre2016lstm} introduces bi-variate Gaussian distribution to sample predictions from the trained mean and variance, which is widely used in recently published methods \cite{shi2021sgcn, Mohamed2020socialstgcnn, zhao2021ExpTraj, bae2022gpgraph, li2024MFAN, xu2024DYNAMICGROUP}. Following this, our BP-SGCN also uses the bi-variate Gaussian distribution to represent the distribution parameters of the predicted trajectories. It outperforms almost all methods under this setting by a significant margin. In addition, both ExpertTraj~\cite{zhao2021ExpTraj} and our BP-SGCN utilize the goal-retrieve mechanism but we have a significant improvement of 10.5\% in ADE and 33.3\% in FDE.}

\RCCC{For generative-based methods, Social-GAN \cite{gupta2018socialgan} is the pioneer method that introduces GANs \cite{goodfellow2014GAN} to generate trajectories with special pooling modules. PECNet \cite{mangalam2020PECNet} utilizes the CVAEs~\cite{sohn2015CVAE} to generate trajectories conditioned on the pre-sampled goal points, which add an extra constraint to the predicted trajectories for better accuracy. Methods like \cite{ye2021agentformer, pei2022socialVAE, Xu2022GroupNetMH, wu2023MSRL, bae2023eigentrajectory, kim2024higher} follow the CVAEs basis to train the encoder with ground-truth trajectories for better latent representations. MID \cite{gu2022stochastic} and LED \cite{mao2023leapfrog} further introduce the diffusion models~\cite{chang2023design} to enhance training and reduce mode collapses. Results reveal that our BP-SGCN outperforms generative-based methods.}

\RCCC{For transformer-based methods, TUTR \cite{shi2023TUTR} proposes a novel global prediction system incorporated with a motion-level transformer encoder and a social-level transformer decoder for accurate trajectory representation. MRGTraj \cite{peng2024mrgtraj} introduces a non-autoregressive enhanced transformer decoder for trajectory prediction. PPT \cite{lin2024progressivePPT} proposes multi-stage transformer progressively modeling trajectories. STGlow \cite{liang2023stglow} further introduces the flow-based generative framework with dual-graphormer to precisely model motion distributions. Compared to STGlow, our BP-SGCN achieves comparable ADE with STGlow, while reducing FDE by 14\%.}

Besides these categories,  LSTM decoder-based methods \cite{wang2023STSF-Net, yang2022GTPPO, sun2021PCCSNet, yang2024TP-EGT} and \cite{wang2023fend} directly predict trajectories using LSTM decoder, which also show comparable results to the transformer-based methods. Social-implicit \cite{mohamed2022socialimplicit} introduces the concept of implicit maximum likelihood estimation mechanism. Memonet~\cite{xu2022memonet}, SICNet\cite{dong2023sparse} and SMEMO \cite{marchetti2024smemo} incorporate memory bank/module concepts into the system, demonstrating considerable performance. Notably, SICNet presents the best results on ETH subset in both ADE and FDE metrics compared with all other methods. Graph-TERN \cite{bae2023TERN} shows a novel trajectory refinement module that first samples the endpoint and then linearly interpolates the predictions. EqMotion \cite{xu2023eqmotion} further introduces the concepts of invariance and equivariance into trajectory prediction to learn motion patterns. Nevertheless, results in \tablename~\ref{tab:ethanducy} illustrate that our BP-SGCN outperforms all of these methods.

\RCCC{For pedestrian-only SDD, \tablename~\ref{tab:sddnolabel} highlights the comparative performance of our BP-SGCN, which secures substantial improvements over all listed models, including the latest STOA models, WTGCN \cite{chen2024wtgcn} and IGGCN \cite{CHEN2025104862}. Specifically, BP-SGCN achieves a 35\% reduction in ADE compared to WTGCN and a 42\% reduction in FDE compared to IGGCN. The results in both heterogeneous SDD and pedestrian-only SDD show the superiority of our BP-SGCN in multiple scenarios.}

\subsection{Qualitative Evaluation}

\begin{figure}[t]
\centering 
     \includegraphics[width=0.40\textwidth]{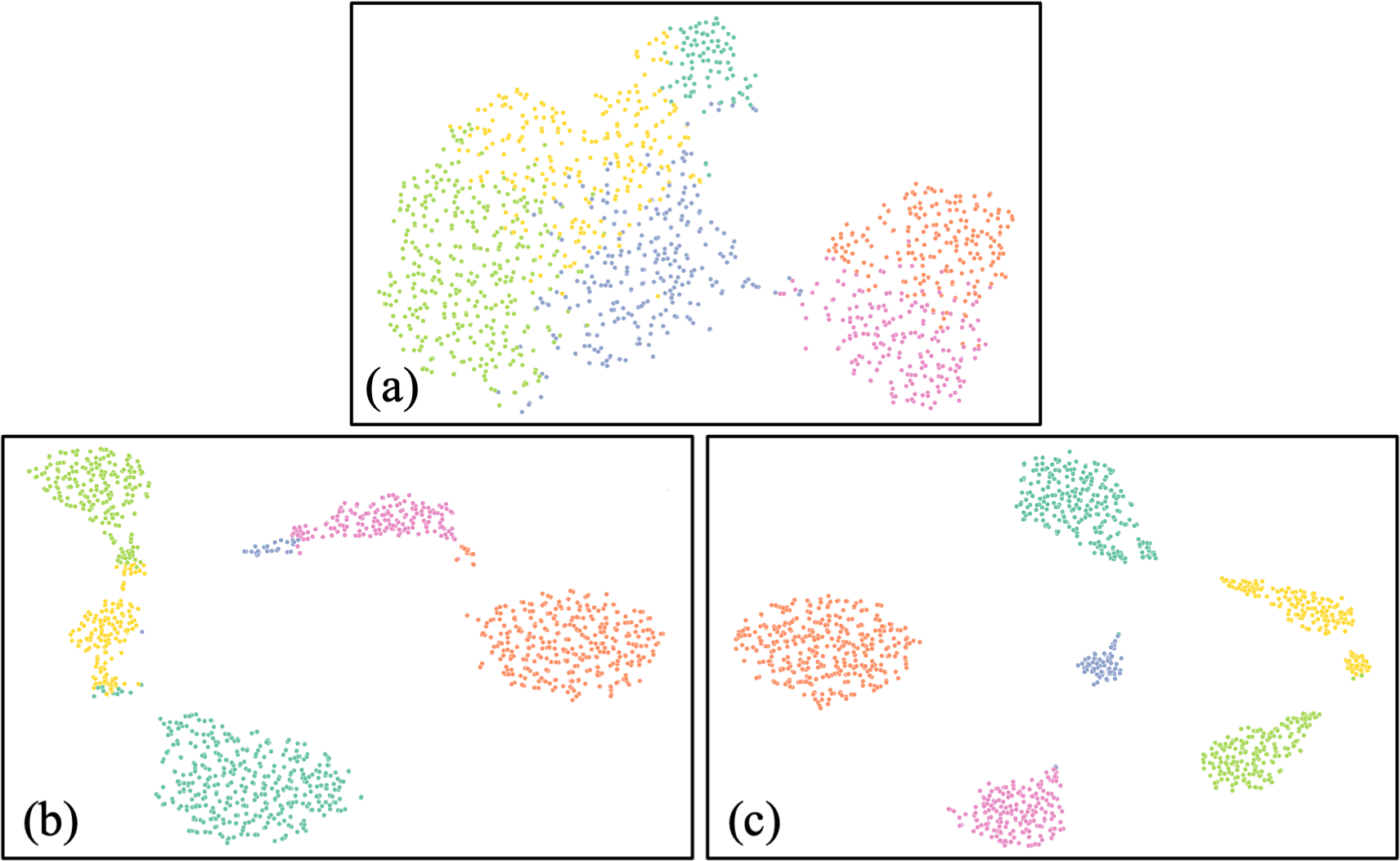}

    \caption{The t-SNE visualization of pseudo-class clustering on SDD ($k$=6) during unsupervised deep clustering. (a) 0 epochs (initialized by k-means), (b) 200 epochs, (c) 800 epochs.} \label{fig:figure3}
    \label{fig:latentvis1}
\end{figure}

\figurename~\ref{fig:latentvis1} shows the t-SNE \cite{laurens2008tsne} visualization of the pseudo-labels in the latent space during unsupervised deep clustering training. At epoch 0, k-means initializes the cluster centers. As the latent is not optimized, class boundaries are unclear. After the training, the VRNN encoder generates a more representative latent for better clustering of the agents' behaviors. 

\begin{figure}[t] \centering
    \includegraphics[width=0.48\textwidth]{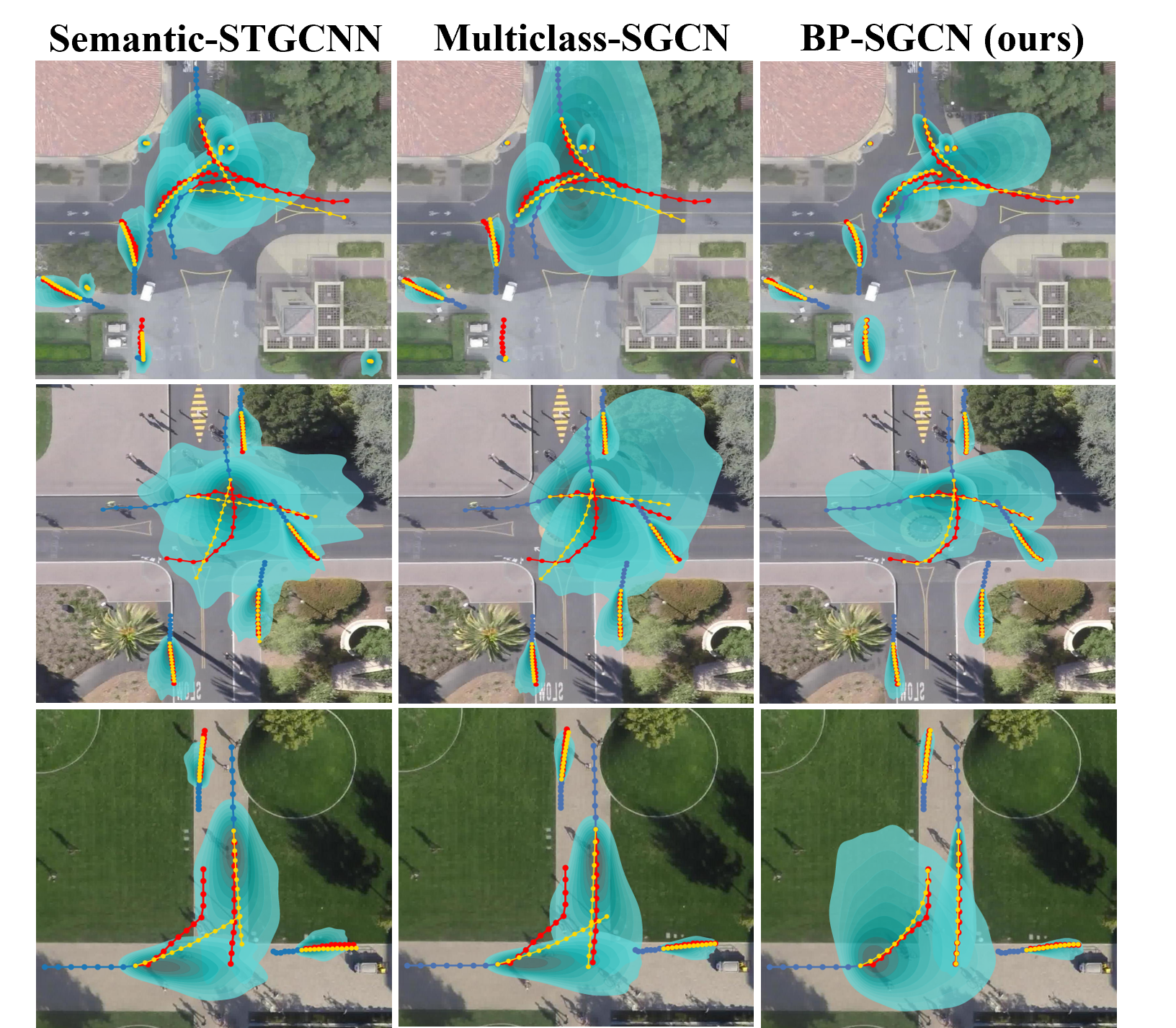}
    \caption{\RCC{Visualization of trajectory prediction on SDD of Semantic-STGCNN \cite{rainbow2021semanticStgcnn}, Multiclass-SGCN\cite{ruochen2022multiclassSGCN}, and BP-SGCN (ours). Blue and red represent observed and ground-truth trajectories respectively, yellow represents the predicted trajectory and light-blue shade represents the predicted distribution.}}
    \label{fig:SDD}
\end{figure}

\begin{figure*}[t]
    \centering
    \includegraphics[width=0.85\textwidth]{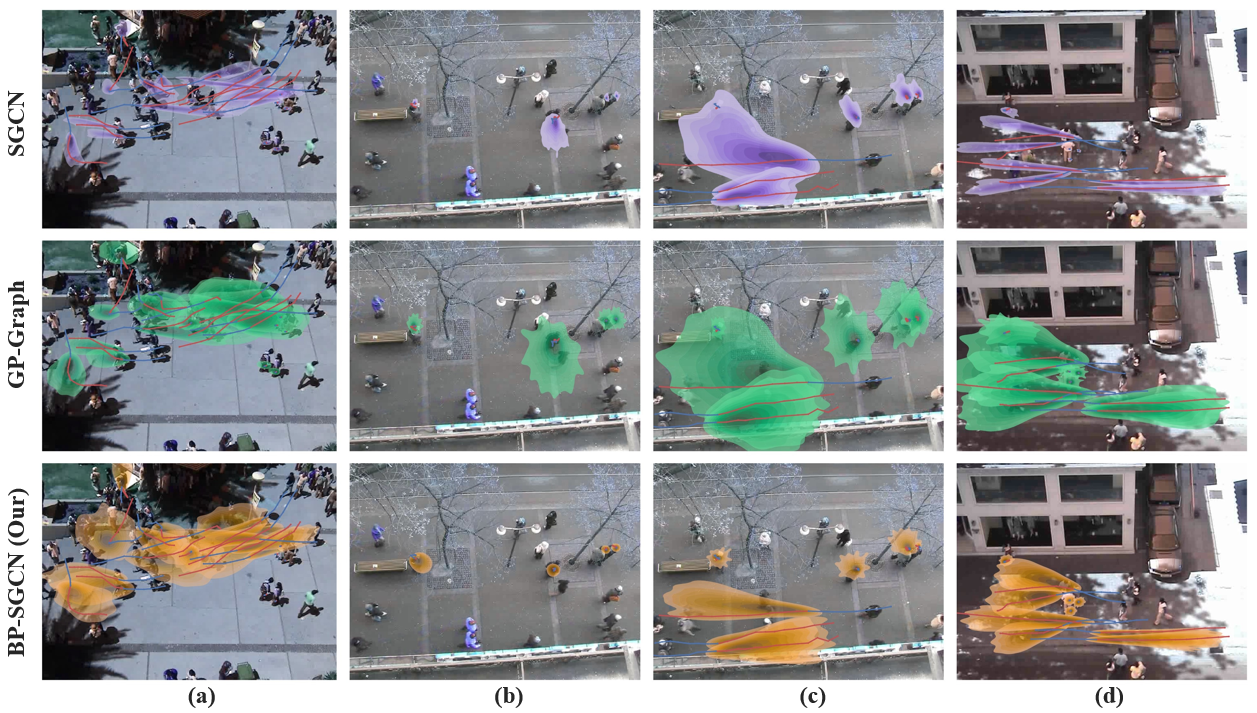}
    \caption{Visualization of the trajectory prediction on ETH/UCY in the scenario of pedestrian walking behaviors. Past trajectories are shown in blue, and ground-truth trajectories are in red. (a) shows the pedestrians in a crowded scenario with complex interactions. (b) shows the scene where four pedestrians are almost static. (c) and (d) show scenes including multiple pedestrian behaviors, such as walking, meeting, and standing.} 
    \label{fig:eth_comparison}
\end{figure*}

\figurename~\ref{fig:SDD} and \figurename~\ref{fig:eth_comparison} visualize the trajectory predictions for the SDD and ETH/UCY datasets, respectively. \RCC{Blue and red dots represent observed and ground-truth future trajectories, respectively. For the SDD dataset, we visualize the predictions in \figurename~\ref{fig:SDD}, where light blue indicates the predicted distributions and yellow dots represent the predicted single trajectory. The visualizations demonstrate that our BP-SGCN exhibits superior performance compared to methods integrating ground-truth labels \cite{ruochen2022multiclassSGCN, rainbow2021semanticStgcnn} in three challenging scenarios characterized by complex social interactions among agents.}

In \figurename~\ref{fig:eth_comparison}, we visualize the predicted distribution in the ETH/UCY datasets across various scenarios, encompassing both simple and complex interactions, and compare our method with SGCN \cite{shi2021sgcn} and GP-Graph \cite{bae2022gpgraph}. We visualize the parameterized distribution of future trajectories, as they are the learning objective of these methods. Qualitative comparisons reveal that our predicted distributions closely align with the ground truth and adeptly capture the non-linear trajectories.
Specifically, scenario (a) illustrates a scene with numerous pedestrians on the street engaging in complex interactions, such as meeting, colliding, and standing still. While all the predicted distributions can accurately represent linear trajectories, both SGCN and GP-Graph falter in predicting the movements of pedestrians exhibiting non-linear behaviors. In contrast, BP-SGCN consistently generates plausible predictions. Scenario (b) displays four stationary pedestrians; however, both SGCN and GP-Graph yield wrong predictions, whereas BP-SGCN accurately captures the static behaviors. In scenario (c), the predicted distributions from both SGCN and GP-Graph demonstrate significant overlaps, leading to a heightened risk of predicted collisions. On the other hand, BP-SGCN's predictions show reduced overlaps. In scenario (d), while GP-Graph continues to display overlap issues, SGCN exhibits overconfidence in its predictions, resulting in a lack of diversity and a propensity to deviate from the ground truth. BP-SGCN effectively addresses both of these challenges, striking a balance between prediction accuracy and diversity.

\subsection{Ablation Study and Parameter Analysis}
\label{sec:results:ablation}

\subsubsection{Cluster Number Analysis}
The effects of cluster number on heterogeneous datasets are shown in \tablename~\ref{tab:sddablation} (Heterogeneous SDD) and \tablename~\ref{tab:argoverse_ablation} (Argoverse 1). The results on pedestrian-only datasets are shown in \tablename~\ref{tab:numclusters} (ETH/UCY) and \tablename~\ref{tab:pedestrian_cluster} (Pedestrian-only SDD). In general, the cluster number depends on the diversity of behaviors, which is strongly correlated with the location. For instance, choosing six clusters for SDD is reasonable given the presence of six types of agents, and this choice yields good performance. Tuning the cluster number for a scene provides extra improvements, and this only has to be done once. These results further reflect that the heterogeneous dataset is more sensitive to the cluster numbers and the pedestrian dataset results exhibit diminished sensitivity, attributable to the inherent behavioral homogeneity and comparatively lower variance observed in human actions. Note that due to its large data size, for Argoverse 1, we run ablation studies and parameter analysis using a partial dataset in a simplified setup.

\RCC{Notably, in our experiments on cluster numbers, a cluster's number equal to 1 denotes that there is no pseudo-label applied on each agent because all the agents are considered to belong to the same class, and consequently the model performance relies solely on the trajectories themselves. In particular, results in \tablename~\ref{tab:numclusters} and \tablename~\ref{tab:pedestrian_cluster} demonstrate that, within datasets exclusively comprising pedestrian agents, our BP-SGCN model is adept at discerning the nuanced variances in their movement patterns. Despite the apparent homogeneity of the agents as pedestrians, our analysis reveals intrinsic behavioral differentiations that our model capitalizes on to significantly improve prediction accuracy. This not only underscores the importance of individualized learning even among seemingly similar entities, but also showcases the efficacy of our model in enhancing predictive outcomes by leveraging these subtle distinctions.}

\begin{table}[t]
\scriptsize
\begin{center}
\renewcommand{\arraystretch}{1.10}
\caption{Cluster number analysis on heterogeneous SDD.}
\label{tab:sddablation}
\begin{tabular}{lcc}
\hline
Clusters & ADE($\downarrow$) & FDE($\downarrow$) \\
\hline\hline
1      & 7.26              &10.03         \\
3      & 7.11              &9.81          \\
6      & \textbf{6.94}     &\textbf{9.57} \\
9      & 7.03              & 9.74         \\
12     & 7.58              & 10.92        \\
\hline
\end{tabular}
\end{center}
\end{table}
\begin{table}[t]
\scriptsize
\begin{center}
\renewcommand{\arraystretch}{1.10}
\caption{Cluster number analysis on Argoverse 1.}
\label{tab:argoverse_ablation}
\begin{tabular}{lcc}
\hline
Clusters & ADE($\downarrow$) & FDE($\downarrow$) \\
\hline\hline
1      & 0.86   &1.63 \\
3      & 0.80 & 1.45 \\
6      & \textbf{0.69} &\textbf{1.15} \\
9      & 0.79 & 1.47 \\
\hline
\end{tabular}
\end{center}

\end{table}

\begin{table}[t]
\scriptsize
  \begin{center}
  \centering
  \renewcommand{\arraystretch}{1.10}
  \caption{Cluster number analysis on ETH/UCY.}
  \label{tab:numclusters}
  \begin{tabular}{l@{}ccccc}
    \hline
    \multirow{2}{*}{Clusters}    & \multicolumn{5}{c}{ADE($\downarrow$) / FDE($\downarrow$)} \\
    &ETH &HOTEL &UNIV &ZARA1 &ZARA2 \\
    \hline\hline
    1 & 0.37/0.51  & 0.14/0.19 & 0.27/0.37 & 0.15/0.21 & 0.24/0.34  \\
    2 & 0.37/0.52  & 0.15/0.21 & 0.18/0.27 & 0.20/0.37 & 0.25/0.35  \\
    3 & 0.45/0.61  & \textbf{0.10}/\textbf{0.14} & 0.27/0.36 & 0.24/0.33 & 0.17/0.34    \\
    4 & \textbf{0.33}/\textbf{0.47}  & 0.12/0.16 & 0.27/0.37 & 0.14/0.20 & \textbf{0.10}/\textbf{0.16}    \\
    5 & 0.36/0.50  & 0.17/0.22& 0.18/0.27 & \textbf{0.13}/\textbf{0.19} & 0.12/0.18    \\

    6 & 0.39/0.53  & 0.15/0.21& 0.18/0.27 & 0.15/0.21 & 0.11/0.17   \\
    7 & 0.37/0.51  & 0.11/0.15& \textbf{0.17}/\textbf{0.26} & 0.26/0.37 & 0.13/0.19 \\
    
    \hline
  \end{tabular}
  \end{center}

\end{table}

\begin{table}[t]
\scriptsize
\begin{center}
\renewcommand{\arraystretch}{1.10}
\caption{Cluster number analysis on pedestrian-only SDD.}
\label{tab:pedestrian_cluster}
\begin{tabular}{lcc}
\hline
Clusters & ADE($\downarrow$) & FDE($\downarrow$) \\
\hline\hline
1      & 0.33   &0.49 \\
3      & \textbf{0.28} & \textbf{0.41} \\
6    & 0.47 & 0.72 \\
9    & 0.31 & 0.47 \\
\hline
\end{tabular}
\end{center}
\end{table}
\subsubsection{Network Components Analysis}
\tablename~\ref{tab:component} shows ablation studies to evidence the effectiveness of network components used in BP-SGCN on heterogeneous and pedestrian-only SDD. The ``No Deep Clustering'' setup uses k-means cluster centers directly for trajectory prediction, and therefore does not implement unsupervised deep learning and end-to-end fine-tuning. The ``No Gumbel-Softmax'' setup directly concatenates the soft assignment to the trajectory features for trajectory prediction. The ``No End-to-End Training'' setup uses only $L_{prediction}$ to optimize the trajectory prediction module but not the deep clustering module; here, the Gumbel-Softmax estimator is substituted with the non-differentiable Argmax function. Results from both the heterogeneous and pedestrian datasets emphasize the significance of all the proposed components in BP-SGCN.

\begin{table}[t]
\scriptsize
\begin{center}
\renewcommand{\arraystretch}{1.10}
\caption{Network components analysis on heterogeneous SDD (upper) and pedestrian-only SDD (lower).}
\label{tab:component}
\begin{tabular}{lcc}
\hline
Method & ADE($\downarrow$) & FDE($\downarrow$) \\
\hline\hline
BP-SGCN (No Deep Clustering)                   & 7.52   & 10.50 \\
BP-SGCN (No Gumbel-Softmax)                    & 7.65   & 10.85 \\
BP-SGCN (No End-to-End Training)               & 10.82  & 15.32 \\
BP-SGCN (Ours)                                 & \textbf{6.94}  &\textbf{9.57} \\
\hline
\end{tabular}
\end{center}
\begin{center}
\renewcommand{\arraystretch}{1.10}
\begin{tabular}{lcc}
\hline

Method & ADE($\downarrow$) & FDE($\downarrow$) \\
\hline\hline
BP-SGCN (No Deep Clustering)                   & 0.30  & 0.44 \\
BP-SGCN (No Gumbel-Softmax)                    & 0.40  & 0.60 \\
BP-SGCN (No End-to-End Training)               & 0.30  & 0.46 \\
BP-SGCN (Ours)                             & \textbf{0.28}  &\textbf{0.41} \\
\hline
\end{tabular}
\end{center}
\end{table}
\RCC{In addition, our proposed Goal-Guided SGCN module utilizes the spatial attention and temporal attention mechanism to enhance the final prediction accuracy. We conduct experiments on ETH/UCY datasets to validate the effectiveness of these two modules. The results shown in \tablename~\ref{ablation_onmodules} indicate that both spatial attention and temporal attention modules are important for the best performance.}

\begin{table}[!t]
\scriptsize
\begin{center}
\renewcommand{\arraystretch}{1.10}
\caption{Prediction module analysis on ETH/UCY datasets.}
\label{ablation_onmodules}
\begin{tabular}{lcc}
\hline
Method & ADE($\downarrow$) & FDE($\downarrow$) \\
\hline\hline
BP-SGCN (No Spatial Attention)                   & 0.25   & 0.30 \\
BP-SGCN (No Temporal Attention)                  & 0.28   & 0.35 \\
BP-SGCN (Ours)                                 & \textbf{0.17}  &\textbf{0.24} \\
\hline
\end{tabular}
\end{center}
\end{table}

\subsubsection{Trajectory Prediction Loss 
Analysis}
As discussed above, we propose a cascaded training strategy with a novel loss function to jointly optimize trajectory prediction and
pseudo-label clustering, defined as:
\begin{equation}
\mathcal{L}_{final} = \mathcal{L}_{prediction} + \mathcal{L}_{cluster}.
\end{equation}

In the proposed loss function, $\mathcal{L}_{prediction}$ and $\mathcal{L}_{cluster}$ contribute equally to the final loss $\mathcal{L}_{final}$. We conduct an ablation study by introducing a weighted sum of losses with a new hyperparameter $\lambda$ to explore the effect and contribution of the two losses on trajectory prediction on both heterogeneous and pedestrian-only SDD datasets:
\begin{equation}
    \mathcal{L}_{final} = \lambda\mathcal{L}_{prediction} + (1 - \lambda)\mathcal{L}_{cluster}.
\end{equation}

Here, we analyze the effect of $\lambda$.
For the proposed BP-SGCN, the default value of $\lambda$ can be considered as 0.5, as both losses contribute equally to the final loss. We further adjust the value of $\lambda$ as 0.25, and 0.75, respectively. The experimental results presented in \tablename~\ref{tab:loss_ablation} show that the performance of BP-SGCN reaches its peak when the ratio of $\mathcal{L}_{prediction}$ and $\mathcal{L}_{cluster}$ is equal, as presented in the main paper, which further indicates that the trajectory prediction and pseudo-label clustering modules are equally important for the overall trajectory prediction performance.

\begin{table}[t]
\scriptsize
\begin{center}
\renewcommand{\arraystretch}{1.10}
\caption{Loss weight analysis between $\mathcal{L}_{prediction}$ and $\mathcal{L}_{cluster}$ on heterogeneous SDD (upper) and pedestrian-only SDD (lower).}
\label{tab:loss_ablation}
\renewcommand{\arraystretch}{1.10}
\begin{tabular}{lcc}
\hline

Method & ADE($\downarrow$) & FDE($\downarrow$) \\
\hline\hline

BP-SGCN ($\lambda = 0.25$)                   & 19.33  & 24.26                \\
BP-SGCN ($\lambda = 0.75$)                   & 7.08  & 9.84                  \\
BP-SGCN (Ours)                               & \textbf{6.94}  &\textbf{9.57} \\
\hline
\end{tabular}
\end{center}

\begin{center}
\renewcommand{\arraystretch}{1.10}
\begin{tabular}{lcc}
\hline

Method & ADE($\downarrow$) & FDE($\downarrow$) \\
\hline\hline
BP-SGCN ($\lambda = 0.25$)                   & 0.46  & 0.70 \\
BP-SGCN ($\lambda = 0.75$)                   & 0.31  & 0.46 \\
BP-SGCN (Ours)                             & \textbf{0.28} &\textbf{0.41} \\
\hline
\end{tabular}
\end{center}

\end{table}


\subsubsection{Clustering Features Analysis}
Finally, \tablename~\ref{tab:featuresforlabel} shows ablation studies on heterogeneous and pedestrian-only SDD datasets with regard to the geometric features used for behavior clustering. These features play a pivotal role, enabling our unsupervised deep clustering module to differentiate agent behaviors effectively. The outcomes highlight the outstanding performance of our proposed features, which integrate relative angle and acceleration magnitude.

\RCC{\subsection{Model Complexity and Inference Time Analysis}}
\label{Sec: Model Complexity and Inference Time Analysis}

\RCCC{To verify the efficiency of our proposed method, we conduct experiments on inference time and model parameters with existing mainstream trajectory prediction frameworks. As demonstrated in \tablename~\ref{tab:complexity}, our method is inferior to EigenTrajectory \cite{bae2023eigentrajectory} and better than all other methods in terms of inference time and model parameters. We leave it as future work to improve the efficiency of our BP-SGCN with more advanced sequential modeling methods such as Transformers \cite{vaswani2017transformer} and State Space Models (SSMs) \cite{gu2021efficiently, gu2023mamba}.}
\begin{table}[H]
\scriptsize
\begin{center}
\renewcommand{\arraystretch}{1.10}
\caption{Clustering features analysis on heterogeneous SDD (upper) and pedestrian-only SDD (lower).}
\label{tab:featuresforlabel}
\begin{tabular}{lcc}
\hline
Method & ADE($\downarrow$) & FDE($\downarrow$) \\
\hline\hline
BP-SGCN (Relative Angle)                        & 19.52 & 34.05 \\
BP-SGCN (Acceleration Magnitude)                & 9.07  & 13.02 \\
BP-SGCN (Ours)                                  & \textbf{6.94}  &\textbf{9.57} \\
\hline
\end{tabular}
\end{center}
\begin{center}
\begin{tabular}{lcc}
\hline
Method & ADE($\downarrow$) & FDE($\downarrow$) \\
\hline\hline
BP-SGCN (Relative Angle)                          & 0.45 & 0.68 \\
BP-SGCN (Acceleration Magnitude)                  & 0.42 & 0.63 \\
BP-SGCN (Ours)                             & \textbf{0.28}  &\textbf{0.41} \\
\hline
\end{tabular}
\end{center}
\end{table}
\begin{table}[!h]
\scriptsize
  \begin{center}
  \renewcommand{\arraystretch}{1.10}
  \setlength{\tabcolsep}{2.5pt}
  \caption{COMPARISON OF THE PROPOSED APPROACHES IN TERMS OF NUMBER OF PARAMETER AND INFERENCE TIME.}
  \label{tab:complexity}
    \begin{tabular}{cccc|rr}
    \hline
    Methods & Venue & Year & & Param \( \times 10^{6} \) & Infer. Time/Iter. \\
    \hline\hline
    ExpertTraj \cite{zhao2021ExpTraj}                   & ICCV      & 2021  && 0.32     & 130 \textit{ms} \\
    Social-VAE \cite{pei2022socialVAE}                  & ECCV      & 2022  && 5.69     & 1110 \textit{ms} \\
    GroupNet \cite{Xu2022GroupNetMH}                    & CVPR      & 2022  && 3.14     & -  \\
    MSRL \cite{wu2023MSRL}                              & AAAI      & 2023  && 11.32    & 970 \textit{ms} \\
    EqMotion \cite{xu2023eqmotion}                      & CVPR      & 2023  && 2.08     & 800 \textit{ms} \\
    TUTR \cite{shi2023TUTR}                             & ICCV      & 2023  && 0.44     & 360 \textit{ms} \\
    EigenTrajectory \cite{bae2023eigentrajectory}       & ICCV      & 2023  && \textbf{0.02}     & \textbf{72} \textit{ms} \\
    \hline
    BP-SGCN (Ours)                                    &           &       && \underline{0.13}    & \underline{110} \textit{ms} \\
    \hline
  \end{tabular}
  \end{center}
\end{table}
\RCC{Moreover, we validate the stability and reliability of our BP-SGCN on heterogeneous trajectory prediction by 10 experiments. Results shown in \tablename~\ref{tab:std_experiment} showcase the stability of our method.}

\begin{table}[!h] 
\centering
\caption{STABILITY TESTS ON ARGOVERSE 1 AND HETEROGENEOUS VERSION OF SDD}
\renewcommand{\arraystretch}{1.10}
\resizebox{0.9\columnwidth}{!}{
    \begin{tabular}{c|*{2}{c}|*{2}{c}}
    \hline
    \multirow{2}{*}{Methods} & \multicolumn{2}{c|}{Argoverse 1} & \multicolumn{2}{c}{SDD}\\
    
    & ADE($\downarrow$) & FDE($\downarrow$) & ADE($\downarrow$) & FDE($\downarrow$) \\
    \hline
    \hline
    BP-SGCN (ours) & $0.68 \pm 0.031$  & $1.16 \pm 0.034$ & $6.97 \pm 0.069$ &$9.59 \pm 0.043$\\
    \hline
    \end{tabular}
}
\label{tab:std_experiment}
\end{table}

\begin{figure*}[t]
    \centering
    \includegraphics[width=0.85\textwidth]{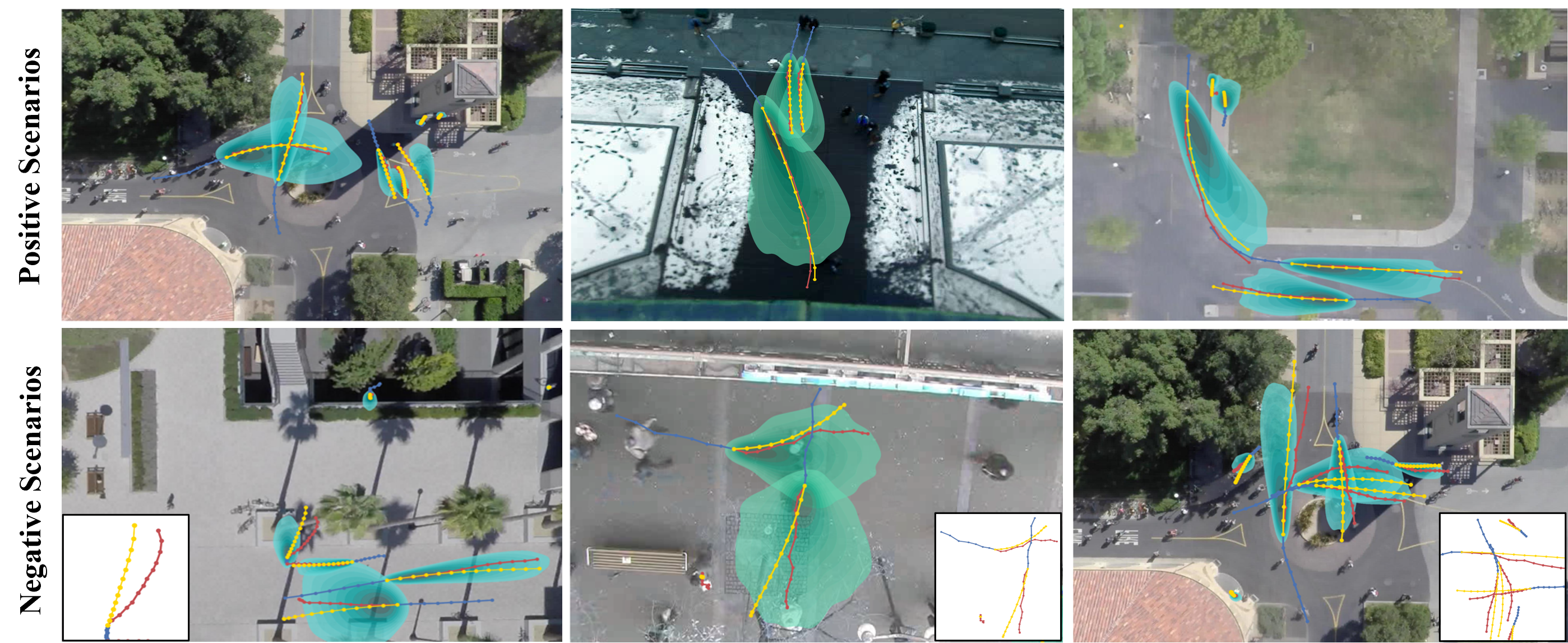}
    \caption{\RCC{Visualization of the trajectory prediction of BP-SGCN in different social scenarios including positive predictions and negative predictions (we highlight erroneous predictions inside the white boxes). Past trajectories are shown in blue, ground-truth trajectories are in red, predicted trajectories are shown in yellow, and distributions are shown in light blue.}}
    \label{fig:pos}
\end{figure*}

\RCC{\subsection{Discussion}}
\label{Sec:Discussion_section}

\RCC{In our experiments, we observed that methods \cite{pei2022socialVAE, wu2023MSRL, bae2023eigentrajectory, shi2023TUTR} tailored exclusively for pedestrians exhibit a sensitivity to the threshold settings that dictate the count of nearby agents. These methods, while ensuring state-of-the-art performance in pedestrian-only trajectory prediction, perform sub-optimally in heterogeneous scenarios due to the challenge of predefining neighbors. The result is shown in \tablename~\ref{tab:pedestrian_heterogeneous}. Unlike these pedestrian-specific approaches, which require manual neighbor selection based on metrics like relative distances, our BP-SGCN model automatically considers all proximate agents as initial neighbors, adaptively filtering out the less relevant ones. Thus, our proposed BP-SGCN is better than these methods in heterogeneous trajectory prediction.}

\begin{table}[H]
\scriptsize
  \begin{center}
  \renewcommand{\arraystretch}{1.10}
  \setlength{\tabcolsep}{2.5pt}
  \caption{RESULTS BY PEDESTRIAN-ONLY METHODS ON THE HETEROGENEOUS VERSION OF SDD.}
  \begin{tabular}{*{2}{c}c|cc}
    \hline
    \label{tab:pedestrian_heterogeneous}
    \multirow{2}{*}{Methods}   &\multirow{2}{*}{Venue} &\multirow{2}{*}{Year} & \multicolumn{2}{c}{ SDD } \\
    &   & &ADE($\downarrow$) &FDE($\downarrow$) \\
    \hline\hline
    Social-VAE + FPC \cite{pei2022socialVAE}                       &ECCV      &2022  &9.41   &\underline{13.49} \\
    MSRL \cite{wu2023MSRL}                                         &AAAI      &2023  &10.72  &16.15 \\
    EigenTrajectory \cite{bae2023eigentrajectory}                  &ICCV      &2023  &\underline{8.85}   &15.15 \\
    TUTR \cite{shi2023TUTR}                                        &ECCV      &2023  &8.93   &15.66 \\
    \hline
    BP-SGCN (Ours)                                              &     &    &\textbf{6.94}    &\textbf{9.57}    \\
    \hline
  \end{tabular}
  \end{center}
\end{table}

\RCC{Next, we showcase inaccurate predictions made by our BP-SGCN and delve into the method's limitations. As depicted in \figurename~\ref{fig:pos}, the first row illustrates the BP-SGCN's proficiency in accurately predicting trajectories across various social contexts. Nonetheless, the second row highlights instances where our BP-SGCN falls short, particularly in scenarios where: 1) trajectories undergo abrupt changes; 2) paths are highly erratic and frequently alter; and 3) social dynamics become exceedingly intricate with numerous agents involved. Looking ahead, our objective is to rectify these inaccuracies by enhancing BP-SGCN's capabilities through the incorporation of cutting-edge deep learning methodologies, including Transformers \cite{vaswani2017transformer} and Diffusion models \cite{mao2023leapfrog}, among others.}

\RCC{The quantity of behavior clusters is an adjustable hyperparameter. We manually select the number of clusters for the unsupervised deep clustering module. This approach brings several challenges, including subjectivity and potential bias, scalability issues, and potential impacts on model performance due to overfitting or underfitting. Moreover, the optimal number of clusters is sensitive to the datasets, which further complicates the selection process. Especially in heterogeneous scenarios, the high variance between different types of agents' motions makes it challenging to identify the best number of clusters to represent behavior features accurately than pedestrian-only scenarios. In the future, we aim to scrutinize the behavior distributions of traffic agents more closely and dynamically estimate the optimal number of clusters \cite{Ronen2022DeepDPMDC, xiao2022deepplugandplay}.}

\RCCC{Despite BP-SGCN’s effectiveness in both heterogeneous and pedestrian-only trajectory prediction, another notable limitation of our model is its current omission of scene semantic features. Although only using trajectories as inputs brings the benefit of computation efficiency and emphasizes the importance of behavior motions, the integration of agent interactions with their surrounding environment can benefit in developing effective trajectory prediction models for use in real-world scenarios\cite{mangalam2021ynet, wong2022V2net, guo2022end2end-grid, lv2023ssagcn}. Recognizing this, a significant direction for our future is to explore how to effectively combine trajectory data with scene semantic features to capture the interactions between static barriers and dynamic agents. We hypothesize that this will not only enhance the model’s prediction accuracy, but also improve the refinement of pseudo-label identification by leveraging the rich context provided by environmental cues.}

\section{Conclusion}
\label{Sec: conclusion}

\RCCC{In this work, we introduce BP-SGCN for heterogeneous and pedestrian trajectory prediction, showcasing its superior performance compared to existing models. In particular, we introduce the concept of behavioral pseudo-labels, which effectively represent the different behavior clusters of agents and do not require extra ground-truth information. BP-SGCN includes a deep unsupervised clustering module that learns the pseudo-label, as well as a pseudo-label informed sparse graph convolution network for trajectory prediction. It implements a cascaded training scheme that first learns the pseudo-labels in an unsupervised manner, and then fine-tunes the labels by optimizing the network end-to-end for better compatibility.}

\RCCC{Beyond pedestrian scenarios, BP-SGCN also shows promising potential in broader domains. In robotic path planning \cite{luo2018porca, liang2023stglow}, BP-SGCN can enhance collision avoidance systems through behavioral pattern analysis \cite{yang2022GTPPO, xue2020poppl} of surrounding agents, facilitating more effective navigation in intricate settings. Additionally, in video monitoring and surveillance systems as suggested in \cite{liang2023stglow, shi2021sgcn}, BP-SGCN can enhance anomaly detection through behavioral pattern analysis of system dynamics, enabling early detection of potential operational irregularities. These applications demonstrate the applicability of BP-SGCN in modeling interactive behaviors across different domains, highlighting its potential for various real-world trajectory prediction tasks.}

\section*{Acknowledgement}
This research is supported in part by the EPSRC NortHFutures project (ref: EP/X031012/1).

\bibliographystyle{IEEEtran}
\bibliography{egbib.bib}

\begin{thebibliography}{100}
\providecommand{\url}[1]{#1}
\csname url@samestyle\endcsname
\providecommand{\newblock}{\relax}
\providecommand{\bibinfo}[2]{#2}
\providecommand{\BIBentrySTDinterwordspacing}{\spaceskip=0pt\relax}
\providecommand{\BIBentryALTinterwordstretchfactor}{4}
\providecommand{\BIBentryALTinterwordspacing}{\spaceskip=\fontdimen2\font plus
\BIBentryALTinterwordstretchfactor\fontdimen3\font minus \fontdimen4\font\relax}
\providecommand{\BIBforeignlanguage}[2]{{%
\expandafter\ifx\csname l@#1\endcsname\relax
\typeout{** WARNING: IEEEtran.bst: No hyphenation pattern has been}%
\typeout{** loaded for the language `#1'. Using the pattern for}%
\typeout{** the default language instead.}%
\else
\language=\csname l@#1\endcsname
\fi
#2}}
\providecommand{\BIBdecl}{\relax}
\BIBdecl

\bibitem{luo2018porca}
Y.~Luo, P.~Cai, A.~Bera, D.~Hsu, W.~S. Lee, and D.~Manocha, ``Porca: Modeling and planning for autonomous driving among many pedestrians,'' \emph{IEEE Robot. Autom. Lett.}, vol.~3, no.~4, pp. 3418--3425, 2018.

\bibitem{yang2024VibrationControl}
W.~Yang, S.~Li, and X.~Luo, ``Data driven vibration control: A review,'' \emph{IEEE/CAA J. Autom. Sin.}, vol.~11, no.~9, pp. 1898--1917, 2024.

\bibitem{joseph2015yolo}
J.~Redmon, S.~Divvala, R.~Girshick, and A.~Farhadi, ``You only look once: Unified, real-time object detection,'' in \emph{Proc. IEEE/CVF Conf. Comput. Vis. Pattern Recognit.}, 2016, pp. 779--788.

\bibitem{kipf2016GCN}
T.~N. Kipf and M.~Welling, ``Semi-supervised classification with graph convolutional networks,'' in \emph{Proc. Int. Conf. Learn. Represent.}, 2017.

\bibitem{Li2024GuiestEditorial}
M.~Li, A.~Micheli, Y.~G. Wang, S.~Pan, P.~Lió, G.~S. Gnecco, and M.~Sanguineti, ``Guest editorial: Deep neural networks for graphs: Theory, models, algorithms, and applications,'' \emph{IEEE Trans. Neural Netw. Learn. Syst.}, vol.~35, no.~4, pp. 4367--4372, 2024.

\bibitem{zheng2022towardGraphSelf}
Y.~Zheng, M.~Jin, S.~Pan, Y.-F. Li, H.~Peng, M.~Li, and Z.~Li, ``Toward graph self-supervised learning with contrastive adjusted zooming,'' \emph{IEEE Trans. Neural Netw. Learn. Syst.}, vol.~35, no.~7, pp. 8882--8896, 2024.

\bibitem{Li2024Heterophilous}
J.~Li, R.~Zheng, H.~Feng, M.~Li, and X.~Zhuang, ``Permutation equivariant graph framelets for heterophilous graph learning,'' \emph{IEEE Trans. Neural Netw. Learn. Syst.}, vol.~35, no.~9, pp. 11\,634--11\,648, 2024.

\bibitem{wu2023GraphIncor}
D.~Wu, Y.~He, and X.~Luo, ``A graph-incorporated latent factor analysis model for high-dimensional and sparse data,'' \emph{IEEE Trans. Emerg. Top. Comput.}, vol.~11, no.~4, pp. 907--917, 2023.

\bibitem{luo2022DirectedNetwork}
X.~Luo, H.~Wu, Z.~Wang, J.~Wang, and D.~Meng, ``A novel approach to large-scale dynamically weighted directed network representation,'' \emph{IEEE Trans. Pattern Anal. Mach. Intell.}, vol.~44, no.~12, pp. 9756--9773, 2022.

\bibitem{shi2021sgcn}
L.~Shi, L.~Wang, C.~Long, S.~Zhou, M.~Zhou, Z.~Niu, and G.~Hua, ``Sgcn: Sparse graph convolution network for pedestrian trajectory prediction,'' in \emph{Proc. IEEE/CVF Conf. Comput. Vis. Pattern Recognit.}, 2021, pp. 8994--9003.

\bibitem{huang2019stgat}
Y.~Huang, H.~Bi, Z.~Li, T.~Mao, and Z.~Wang, ``Stgat: Modeling spatial-temporal interactions for human trajectory prediction,'' in \emph{Proc. IEEE/CVF Int. Conf. Comput. Vis.}, 2019, pp. 6272--6281.

\bibitem{Mohamed2020socialstgcnn}
A.~Mohamed, K.~Qian, M.~Elhoseiny, and C.~Claudel, ``Social-stgcnn: A social spatio-temporal graph convolutional neural network for human trajectory prediction,'' in \emph{Proc. IEEE/CVF Conf. Comput. Vis. Pattern Recognit.}, 2020, pp. 14\,424--14\,432.

\bibitem{mohamed2022socialimplicit}
A.~Mohamed, D.~Zhu, W.~Vu, M.~Elhoseiny, and C.~Claudel, ``Social-implicit: Rethinking trajectory prediction evaluation and the effectiveness of implicit maximum likelihood estimation,'' in \emph{Proc. Eur. Conf. Comput. Vis.}, 2022, pp. 463--479.

\bibitem{xu2023DecoupleSqueeze}
B.~Xu, X.~Shu, J.~Zhang, G.~Dai, and Y.~Song, ``Spatiotemporal decouple-and-squeeze contrastive learning for semisupervised skeleton-based action recognition,'' \emph{IEEE Trans. Neural Netw. Learn. Syst.}, 2023.

\bibitem{shu2022anchor-contrastive}
X.~Shu, B.~Xu, L.~Zhang, and J.~Tang, ``Multi-granularity anchor-contrastive representation learning for semi-supervised skeleton-based action recognition,'' \emph{IEEE Trans. Pattern Anal. Mach. Intell.}, 2022.

\bibitem{xu2022xinvariant}
B.~Xu, X.~Shu, and Y.~Song, ``X-invariant contrastive augmentation and representation learning for semi-supervised skeleton-based action recognition,'' \emph{IEEE Trans. Image Process.}, vol.~31, pp. 3852--3867, 2022.

\bibitem{xu2023polymerization}
B.~Xu and X.~Shu, ``Pyramid self-attention polymerization learning for semi-supervised skeleton-based action recognition,'' \emph{arXiv preprint arXiv:2302.02327}, 2023.

\bibitem{qiao20222ggcn}
T.~Qiao, Q.~Men, F.~W.~B. Li, Y.~Kubotani, S.~Morishima, and H.~P.~H. Shum, ``Geometric features informed multi-person human-object interaction recognition in videos,'' in \emph{Proc. Eur. Conf. Comput. Vis.}, 2022.

\bibitem{sun2021PCCSNet}
J.~Sun, Y.~Li, H.-S. Fang, and C.~Lu, ``Three steps to multimodal trajectory prediction: Modality clustering, classification and synthesis,'' in \emph{Proc. IEEE/CVF Int. Conf. Comput. Vis.}, 2021, pp. 13\,250--13\,259.

\bibitem{pellegrini2009ETH}
S.~Pellegrini, A.~Ess, K.~Schindler, and L.~Van~Gool, ``You'll never walk alone: Modeling social behavior for multi-target tracking,'' in \emph{Proc. IEEE/CVF Int. Conf. Comput. Vis.}\hskip 1em plus 0.5em minus 0.4em\relax IEEE, 2009, pp. 261--268.

\bibitem{Lerner2007UCY}
A.~Lerner, Y.~Chrysanthou, and D.~Lischinski, ``Crowds by example,'' in \emph{Computer graphics forum}, vol.~26, no.~3.\hskip 1em plus 0.5em minus 0.4em\relax Wiley Online Library, 2007, pp. 655--664.

\bibitem{mangalam2021ynet}
K.~Mangalam, Y.~An, H.~Girase, and J.~Malik, ``From goals, waypoints \& paths to long term human trajectory forecasting,'' in \emph{Proc. IEEE/CVF Int. Conf. Comput. Vis.}, 2021, pp. 15\,233--15\,242.

\bibitem{zhao2021ExpTraj}
H.~Zhao and R.~P. Wildes, ``Where are you heading? dynamic trajectory prediction with expert goal examples,'' in \emph{Proc. IEEE/CVF Int. Conf. Comput. Vis.}, 2021, pp. 7629--7638.

\bibitem{mangalam2020PECNet}
K.~Mangalam, H.~Girase, S.~Agarwal, K.-H. Lee, E.~Adeli, J.~Malik, and A.~Gaidon, ``It is not the journey but the destination: Endpoint conditioned trajectory prediction,'' in \emph{Proc. Eur. Conf. Comput. Vis.}, 2020, pp. 759--776.

\bibitem{Alexandre2016lstm}
A.~Alahi, K.~Goel, V.~Ramanathan, A.~Robicquet, L.~Fei-Fei, and S.~Savarese, ``Social lstm: Human trajectory prediction in crowded spaces,'' in \emph{Proc. IEEE/CVF Conf. Comput. Vis. Pattern Recognit.}, 2016, pp. 961--971.

\bibitem{rainbow2021semanticStgcnn}
B.~A. Rainbow, Q.~Men, and H.~P. Shum, ``Semantics-stgcnn: A semantics-guided spatial-temporal graph convolutional network for multi-class trajectory prediction,'' in \emph{IEEE Int. Conf. Syst. Man Cybern.}\hskip 1em plus 0.5em minus 0.4em\relax IEEE, 2021, pp. 2959--2966.

\bibitem{fang2021UNIN}
F.~Zheng, L.~Wang, S.~Zhou, W.~Tang, Z.~Niu, N.~Zheng, and G.~Hua, ``Unlimited neighborhood interaction for heterogeneous trajectory prediction,'' in \emph{Proc. IEEE/CVF Int. Conf. Comput. Vis.}, 2021, pp. 13\,168--13\,177.

\bibitem{ruochen2022multiclassSGCN}
R.~Li, S.~Katsigiannis, and H.~P. Shum, ``Multiclass-sgcn: Sparse graph-based trajectory prediction with agent class embedding,'' in \emph{IEEE Int. Conf. Image Process.}\hskip 1em plus 0.5em minus 0.4em\relax IEEE, 2022, pp. 2346--2350.

\bibitem{du2024SFEM-GCN}
Q.~Du, X.~Wang, S.~Yin, L.~Li, and H.~Ning, ``Social force embedded mixed graph convolutional network for multi-class trajectory prediction,'' \emph{IEEE Trans. Intell. Veh.}, pp. 1--11, 2024.

\bibitem{Robicquet2016SDD}
A.~Robicquet, A.~Sadeghian, A.~Alahi, and S.~Savarese, ``Learning social etiquette: Human trajectory understanding in crowded scenes,'' in \emph{Proc. Eur. Conf. Comput. Vis.}, 2016, pp. 549--565.

\bibitem{junyuan2015DEC}
J.~Xie, R.~Girshick, and A.~Farhadi, ``Unsupervised deep embedding for clustering analysis,'' in \emph{Proc. Int. Conf. Mach. Learn.}, 2016, pp. 478--487.

\bibitem{Junyoung2015VRNN}
J.~Chung, K.~Kastner, L.~Dinh, K.~Goel, A.~C. Courville, and Y.~Bengio, ``A recurrent latent variable model for sequential data,'' \emph{Proc. Adv. Neu. Inf. Process. Syst.}, vol.~28, 2015.

\bibitem{yang2023ManiCalibration}
W.~Yang, S.~Li, Z.~Li, and X.~Luo, ``Highly accurate manipulator calibration via extended kalman filter-incorporated residual neural network,'' \emph{IEEE Trans. Ind. Inform.}, vol.~19, no.~11, pp. 10\,831--10\,841, 2023.

\bibitem{mingfang2019argoverse}
M.-F. Chang, J.~Lambert, P.~Sangkloy, J.~Singh, S.~Bak, A.~Hartnett, D.~Wang, P.~Carr, S.~Lucey, D.~Ramanan \emph{et~al.}, ``Argoverse: 3d tracking and forecasting with rich maps,'' in \emph{Proc. IEEE/CVF Conf. Comput. Vis. Pattern Recognit.}, 2019, pp. 8748--8757.

\bibitem{Becker2018AnEO}
S.~Becker, R.~Hug, W.~H{\"u}bner, and M.~Arens, ``An evaluation of trajectory prediction approaches and notes on the trajnet benchmark,'' \emph{arXiv preprint arXiv:1805.07663}, 2018.

\bibitem{HochSchm1997lstm}
S.~Hochreiter and J.~Schmidhuber, ``Long short-term memory,'' \emph{Neural computation}, vol.~9, no.~8, pp. 1735--1780, 1997.

\bibitem{gupta2018socialgan}
A.~Gupta, J.~Johnson, L.~Fei-Fei, S.~Savarese, and A.~Alahi, ``Social gan: Socially acceptable trajectories with generative adversarial networks,'' in \emph{Proc. IEEE/CVF Conf. Comput. Vis. Pattern Recognit.}, 2018, pp. 2255--2264.

\bibitem{goodfellow2014GAN}
I.~Goodfellow, J.~Pouget-Abadie, M.~Mirza, B.~Xu, D.~Warde-Farley, S.~Ozair, A.~Courville, and Y.~Bengio, ``Generative adversarial nets,'' \emph{Proc. Adv. Neu. Inf. Process. Syst.}, vol.~27, 2014.

\bibitem{chang2023design}
Z.~Chang, G.~A. Koulieris, and H.~P.~H. Shum, ``On the design fundamentals of diffusion models: A survey,'' \emph{arXiv}, 2023.

\bibitem{gu2022stochastic}
T.~Gu, G.~Chen, J.~Li, C.~Lin, Y.~Rao, J.~Zhou, and J.~Lu, ``Stochastic trajectory prediction via motion indeterminacy diffusion,'' in \emph{Proc. IEEE/CVF Conf. Comput. Vis. Pattern Recognit.}, 2022, pp. 17\,113--17\,122.

\bibitem{mao2023leapfrog}
W.~Mao, C.~Xu, Q.~Zhu, S.~Chen, and Y.~Wang, ``Leapfrog diffusion model for stochastic trajectory prediction,'' in \emph{Proc. IEEE/CVF Conf. Comput. Vis. Pattern Recognit.}, 2023, pp. 5517--5526.

\bibitem{ma2019trafficPredict}
Y.~Ma, X.~Zhu, S.~Zhang, R.~Yang, W.~Wang, and D.~Manocha, ``Trafficpredict: Trajectory prediction for heterogeneous traffic-agents,'' in \emph{Proc. AAAI Conf. Art. Intel.}, vol.~33, no.~01, 2019, pp. 6120--6127.

\bibitem{cunjun2020star}
C.~Yu, X.~Ma, J.~Ren, H.~Zhao, and S.~Yi, ``Spatio-temporal graph transformer networks for pedestrian trajectory prediction,'' in \emph{Proc. Eur. Conf. Comput. Vis.}, 2020, pp. 507--523.

\bibitem{vaswani2017transformer}
A.~Vaswani, N.~Shazeer, N.~Parmar, J.~Uszkoreit, L.~Jones, A.~N. Gomez, {\L}.~Kaiser, and I.~Polosukhin, ``Attention is all you need,'' \emph{Proc. Adv. Neu. Inf. Process. Syst.}, vol.~30, 2017.

\bibitem{ye2021agentformer}
Y.~Yuan, X.~Weng, Y.~Ou, and K.~M. Kitani, ``Agentformer: Agent-aware transformers for socio-temporal multi-agent forecasting,'' in \emph{Proc. IEEE/CVF Int. Conf. Comput. Vis.}, 2021, pp. 9813--9823.

\bibitem{Xu2022GroupNetMH}
C.~Xu, M.~Li, Z.~Ni, Y.~Zhang, and S.~Chen, ``Groupnet: Multiscale hypergraph neural networks for trajectory prediction with relational reasoning,'' \emph{Proc. IEEE/CVF Conf. Comput. Vis. Pattern Recognit.}, pp. 6488--6497, 2022.

\bibitem{pei2022socialVAE}
P.~Xu, J.-B. Hayet, and I.~Karamouzas, ``Socialvae: Human trajectory prediction using timewise latents,'' in \emph{Proc. Eur. Conf. Comput. Vis.}, 2022, pp. 511--528.

\bibitem{xu2022memonet}
C.~Xu, W.~Mao, W.~Zhang, and S.~Chen, ``Remember intentions: Retrospective-memory-based trajectory prediction,'' in \emph{Proc. IEEE/CVF Conf. Comput. Vis. Pattern Recognit.}, 2022, pp. 6488--6497.

\bibitem{li2024GoalOriented}
D.~Li, Q.~Zhang, S.~Lu, Y.~Pan, and D.~Zhao, ``Conditional goal-oriented trajectory prediction for interacting vehicles,'' \emph{IEEE Trans. Neural Netw. Learn. Syst.}, vol.~35, no.~12, pp. 18\,758--18\,770, 2024.

\bibitem{sohn2015CVAE}
K.~Sohn, H.~Lee, and X.~Yan, ``Learning structured output representation using deep conditional generative models,'' \emph{Proc. Adv. Neu. Inf. Process. Syst.}, vol.~28, 2015.

\bibitem{lee2017desire}
N.~Lee, W.~Choi, P.~Vernaza, C.~B. Choy, P.~H. Torr, and M.~Chandraker, ``Desire: Distant future prediction in dynamic scenes with interacting agents,'' in \emph{Proc. IEEE/CVF Conf. Comput. Vis. Pattern Recognit.}, 2017, pp. 336--345.

\bibitem{wong2022V2net}
C.~Wong, B.~Xia, Z.~Hong, Q.~Peng, W.~Yuan, Q.~Cao, Y.~Yang, and X.~You, ``View vertically: A hierarchical network for trajectory prediction via fourier spectrums,'' in \emph{Proc. Eur. Conf. Comput. Vis.}, 2022, pp. 682--700.

\bibitem{fang2023HSG}
J.~Fang, C.~Zhu, P.~Zhang, H.~Yu, and J.~Xue, ``Heterogeneous trajectory forecasting via risk and scene graph learning,'' \emph{IEEE Trans. Intell. Transp. Syst.}, 2023.

\bibitem{guo2022end2end-grid}
K.~Guo, W.~Liu, and J.~Pan, ``End-to-end trajectory distribution prediction based on occupancy grid maps,'' in \emph{Proc. IEEE/CVF Conf. Comput. Vis. Pattern Recognit.}, 2022, pp. 2242--2251.

\bibitem{huikun2019JPKT}
H.~Bi, Z.~Fang, T.~Mao, Z.~Wang, and Z.~Deng, ``Joint prediction for kinematic trajectories in vehicle-pedestrian-mixed scenes,'' in \emph{Proc. IEEE/CVF Int. Conf. Comput. Vis.}, 2019, pp. 10\,383--10\,392.

\bibitem{Tung2020covernet}
T.~Phan-Minh, E.~C. Grigore, F.~A. Boulton, O.~Beijbom, and E.~M. Wolff, ``Covernet: Multimodal behavior prediction using trajectory sets,'' in \emph{Proc. IEEE/CVF Conf. Comput. Vis. Pattern Recognit.}, 2020, pp. 14\,074--14\,083.

\bibitem{chen2024AEMTP}
S.~Chen and J.~Wang, ``Heterogeneous interaction modeling with reduced accumulated error for multiagent trajectory prediction,'' \emph{IEEE Trans. Neural Netw. Learn. Syst.}, vol.~35, no.~6, pp. 8040--8052, 2024.

\bibitem{zhang2024SMGCN}
J.~Zhang, J.~Yao, L.~Yan, Y.~Xu, and Z.~Wang, ``Sparse multi-relational graph convolutional network for multi-type object trajectory prediction,'' in \emph{Int. Joint Conf. Artif. Intell.}, 2024, pp. 1697--1705.

\bibitem{fernando2018soft+}
T.~Fernando, S.~Denman, S.~Sridharan, and C.~Fookes, ``Soft+ hardwired attention: An lstm framework for human trajectory prediction and abnormal event detection,'' \emph{Neural networks}, vol. 108, pp. 466--478, 2018.

\bibitem{xue2020poppl}
H.~Xue, D.~Q. Huynh, and M.~Reynolds, ``Poppl: Pedestrian trajectory prediction by lstm with automatic route class clustering,'' \emph{IEEE Trans. Neural Netw. Learn. Syst.}, vol.~32, no.~1, pp. 77--90, 2020.

\bibitem{lawal2016support}
I.~A. Lawal, F.~Poiesi, D.~Anguita, and A.~Cavallaro, ``Support vector motion clustering,'' \emph{IEEE Trans. Circuits Syst. Video Technol.}, vol.~27, no.~11, pp. 2395--2408, 2016.

\bibitem{lui2021DDPTP}
A.~K.-F. Lui, Y.-H. Chan, and M.-F. Leung, ``Modelling of destinations for data-driven pedestrian trajectory prediction in public buildings,'' in \emph{IEEE Int. Conf. Big Data}.\hskip 1em plus 0.5em minus 0.4em\relax IEEE, 2021, pp. 1709--1717.

\bibitem{bae2022gpgraph}
I.~Bae, J.-H. Park, and H.-G. Jeon, ``Learning pedestrian group representations for multi-modal trajectory prediction,'' in \emph{Proc. Eur. Conf. Comput. Vis.}, 2022, pp. 270--289.

\bibitem{wang2023fend}
Y.~Wang, P.~Zhang, L.~Bai, and J.~Xue, ``Fend: A future enhanced distribution-aware contrastive learning framework for long-tail trajectory prediction,'' in \emph{Proc. IEEE/CVF Conf. Comput. Vis. Pattern Recognit.}, 2023, pp. 1400--1409.

\bibitem{laurens2008tsne}
L.~Van~der Maaten and G.~Hinton, ``Visualizing data using t-sne.'' \emph{J. Mach. Learn. Res.}, vol.~9, no.~11, 2008.

\bibitem{sai2018DTC}
N.~Sai~Madiraju, S.~M. Sadat, D.~Fisher, and H.~Karimabadi, ``Deep temporal clustering: Fully unsupervised learning of time-domain features,'' \emph{arXiv e-prints}, pp. arXiv--1802, 2018.

\bibitem{kingma2013VAE}
D.~P. Kingma and M.~Welling, ``Auto-encoding variational bayes,'' \emph{arXiv preprint arXiv:1312.6114}, 2013.

\bibitem{Junyoung2014GRU}
J.~Chung, C.~Gulcehre, K.~Cho, and Y.~Bengio, ``Empirical evaluation of gated recurrent neural networks on sequence modeling,'' \emph{arXiv preprint arXiv:1412.3555}, 2014.

\bibitem{cuturi2017softdtw}
M.~Cuturi and M.~Blondel, ``Soft-dtw: a differentiable loss function for time-series,'' in \emph{Proc. Int. Conf. Mach. Learn.}, 2017, pp. 894--903.

\bibitem{sakoe1978DTW}
H.~Sakoe and S.~Chiba, ``Dynamic programming algorithm optimization for spoken word recognition,'' \emph{IEEE transactions on acoustics, speech, and signal processing}, vol.~26, no.~1, pp. 43--49, 1978.

\bibitem{zhao2018shapeDTW}
J.~Zhao and L.~Itti, ``shapedtw: Shape dynamic time warping,'' \emph{Pattern Recognition}, vol.~74, pp. 171--184, 2018.

\bibitem{MacQueen1967kmeans}
J.~MacQueen \emph{et~al.}, ``Some methods for classification and analysis of multivariate observations,'' in \emph{Proc. Fifth Berkeley Symp. Math. Statist. Probab.}, vol.~1, no.~14, 1967, pp. 281--297.

\bibitem{Reynolds2009GMM}
D.~A. Reynolds \emph{et~al.}, ``Gaussian mixture models.'' \emph{Encyclopedia of biometrics}, vol. 741, no. 659-663, 2009.

\bibitem{Jang2016gumbelsoftmax}
E.~Jang, S.~Gu, and B.~Poole, ``Categorical reparameterization with gumbel-softmax,'' \emph{arXiv preprint arXiv:1611.01144}, 2016.

\bibitem{li2021temporalpyramid}
Y.~Li, R.~Liang, W.~Wei, W.~Wang, J.~Zhou, and X.~Li, ``Temporal pyramid network with spatial-temporal attention for pedestrian trajectory prediction,'' \emph{IEEE Trans. Netw. Sci. Eng.}, vol.~9, no.~3, pp. 1006--1019, 2021.

\bibitem{junwei2020multiverse}
J.~Liang, L.~Jiang, K.~Murphy, T.~Yu, and A.~Hauptmann, ``The garden of forking paths: Towards multi-future trajectory prediction,'' in \emph{Proc. IEEE/CVF Conf. Comput. Vis. Pattern Recognit.}, 2020, pp. 10\,508--10\,518.

\bibitem{zhang2023bip-tree}
Y.~Zhang, W.~Guo, J.~Su, P.~Lv, and M.~Xu, ``Bip-tree: Tree variant with behavioral intention perception for heterogeneous trajectory prediction,'' \emph{IEEE Trans. Intell. Transp. Syst.}, 2023.

\bibitem{zhou2022Hivt}
Z.~Zhou, L.~Ye, J.~Wang, K.~Wu, and K.~Lu, ``Hivt: Hierarchical vector transformer for multi-agent motion prediction,'' in \emph{Proc. IEEE/CVF Conf. Comput. Vis. Pattern Recognit.}, 2022, pp. 8813--8823.

\bibitem{monti2021dagnet}
A.~Monti, A.~Bertugli, S.~Calderara, and R.~Cucchiara, ``Dag-net: Double attentive graph neural network for trajectory forecasting,'' in \emph{Int. Conf. Pattern Recognit.}\hskip 1em plus 0.5em minus 0.4em\relax IEEE, 2021, pp. 2551--2558.

\bibitem{tianyang2019MATF}
T.~Zhao, Y.~Xu, M.~Monfort, W.~Choi, C.~Baker, Y.~Zhao, Y.~Wang, and Y.~N. Wu, ``Multi-agent tensor fusion for contextual trajectory prediction,'' in \emph{Proc. IEEE/CVF Conf. Comput. Vis. Pattern Recognit.}, 2019, pp. 12\,126--12\,134.

\bibitem{liang2020simaug}
J.~Liang, L.~Jiang, and A.~Hauptmann, ``Simaug: Learning robust representations from simulation for trajectory prediction,'' in \emph{Proc. Eur. Conf. Comput. Vis.}, 2020, pp. 275--292.

\bibitem{wang2023STSF-Net}
Y.~Wang and S.~Chen, ``Multi-agent trajectory prediction with spatio-temporal sequence fusion,'' \emph{IEEE Trans. Multimedia}, 2021.

\bibitem{azadani2023capha}
M.~N. Azadani and A.~Boukerche, ``Capha: A novel context-aware behavior prediction system of heterogeneous agents for autonomous vehicles,'' \emph{IEEE Trans. Veh. Technol}, 2023.

\bibitem{chen2023NVAGT}
X.~Chen, H.~Zhang, Y.~Hu, J.~Liang, and H.~Wang, ``Vnagt: Variational non-autoregressive graph transformer network for multi-agent trajectory prediction,'' \emph{IEEE Trans. Veh. Technol}, vol.~72, no.~10, pp. 12\,540--12\,552, 2023.

\bibitem{Rhinehart2018r2p2AR}
N.~Rhinehart, K.~M. Kitani, and P.~Vernaza, ``R2p2: A reparameterized pushforward policy for diverse, precise generative path forecasting,'' in \emph{Proc. Eur. Conf. Comput. Vis.}, 2018, pp. 772--788.

\bibitem{Park2020DiverseAA}
S.~H. Park, G.~Lee, J.~Seo, M.~Bhat, M.~Kang, J.~Francis, A.~Jadhav, P.~P. Liang, and L.-P. Morency, ``Diverse and admissible trajectory forecasting through multimodal context understanding,'' in \emph{Proc. Eur. Conf. Comput. Vis.}, 2020, pp. 282--298.

\bibitem{Tang2019MultipleFP}
C.~Tang and R.~R. Salakhutdinov, ``Multiple futures prediction,'' \emph{Proc. Adv. Neu. Inf. Process. Syst.}, vol.~32, 2019.

\bibitem{xu2022DD}
X.~Xu, W.~Liu, and L.~Yu, ``Trajectory prediction for heterogeneous traffic-agents using knowledge correction data-driven model,'' \emph{Information Sciences}, vol. 608, pp. 375--391, 2022.

\bibitem{yang2022GTPPO}
B.~Yang, G.~Yan, P.~Wang, C.-Y. Chan, X.~Song, and Y.~Chen, ``A novel graph-based trajectory predictor with pseudo-oracle,'' \emph{IEEE Trans. Neural Netw. Learn. Syst.}, vol.~33, no.~12, pp. 7064--7078, 2021.

\bibitem{bae2023TERN}
I.~Bae and H.-G. Jeon, ``A set of control points conditioned pedestrian trajectory prediction,'' in \emph{Proc. AAAI Conf. Art. Intel.}, vol.~37, no.~5, 2023, pp. 6155--6165.

\bibitem{wu2023MSRL}
Y.~Wu, L.~Wang, S.~Zhou, J.~Duan, G.~Hua, and W.~Tang, ``Multi-stream representation learning for pedestrian trajectory prediction,'' in \emph{Proc. AAAI Conf. Art. Intel.}, vol.~37, no.~3, 2023, pp. 2875--2882.

\bibitem{xu2023eqmotion}
C.~Xu, R.~T. Tan, Y.~Tan, S.~Chen, Y.~G. Wang, X.~Wang, and Y.~Wang, ``Eqmotion: Equivariant multi-agent motion prediction with invariant interaction reasoning,'' in \emph{Proc. IEEE/CVF Conf. Comput. Vis. Pattern Recognit.}, 2023, pp. 1410--1420.

\bibitem{bae2023eigentrajectory}
I.~Bae, J.~Oh, and H.-G. Jeon, ``Eigentrajectory: Low-rank descriptors for multi-modal trajectory forecasting,'' \emph{arXiv preprint arXiv:2307.09306}, 2023.

\bibitem{shi2023TUTR}
L.~Shi, L.~Wang, S.~Zhou, and G.~Hua, ``Trajectory unified transformer for pedestrian trajectory prediction,'' in \emph{Proc. IEEE/CVF Int. Conf. Comput. Vis.}, 2023, pp. 9675--9684.

\bibitem{dong2023sparse}
Y.~Dong, L.~Wang, S.~Zhou, and G.~Hua, ``Sparse instance conditioned multimodal trajectory prediction,'' in \emph{Proc. IEEE/CVF Int. Conf. Comput. Vis.}, 2023, pp. 9763--9772.

\bibitem{yang2024TP-EGT}
B.~Yang, F.~Fan, R.~Ni, H.~Wang, A.~Jafaripournimchahi, and H.~Hu, ``A multi-task learning network with a collision-aware graph transformer for traffic-agents trajectory prediction,'' \emph{IEEE Trans. Intell. Transp. Syst.}, 2024.

\bibitem{xu2024DYNAMICGROUP}
C.~Xu, Y.~Wei, B.~Tang, S.~Yin, Y.~Zhang, S.~Chen, and Y.~Wang, ``Dynamic-group-aware networks for multi-agent trajectory prediction with relational reasoning,'' \emph{Neural Networks}, vol. 170, pp. 564--577, 2024.

\bibitem{marchetti2024smemo}
F.~Marchetti, F.~Becattini, L.~Seidenari, and A.~Del~Bimbo, ``Smemo: social memory for trajectory forecasting,'' \emph{IEEE Trans. Pattern Anal. Mach. Intell.}, 2024.

\bibitem{liang2023stglow}
R.~Liang, Y.~Li, J.~Zhou, and X.~Li, ``Stglow: A flow-based generative framework with dual-graphormer for pedestrian trajectory prediction,'' \emph{IEEE Trans. Neural Netw. Learn. Syst.}, vol.~35, no.~11, pp. 16\,504--16\,517, 2024.

\bibitem{peng2024mrgtraj}
Y.~Peng, G.~Zhang, J.~Shi, X.~Li, and L.~Zheng, ``Mrgtraj: A novel non-autoregressive approach for human trajectory prediction,'' \emph{IEEE Trans. Circuits Syst. Video Technol.}, vol.~34, no.~4, pp. 2318--2331, 2024.

\bibitem{kim2024higher}
S.~Kim, H.-g. Chi, H.~Lim, K.~Ramani, J.~Kim, and S.~Kim, ``Higher-order relational reasoning for pedestrian trajectory prediction,'' in \emph{Proc. IEEE Conf. Comput. Vis. Pattern Recognit.}, 2024, pp. 15\,251--15\,260.

\bibitem{lin2024progressivePPT}
X.~Lin, T.~Liang, J.~Lai, and J.-F. Hu, ``Progressive pretext task learning for human trajectory prediction,'' in \emph{Proc. Eur. Conf. Comput. Vis.}, 2024, pp. 197--214.

\bibitem{amirian2019socialways}
J.~Amirian, J.-B. Hayet, and J.~Pettr{\'e}, ``Social ways: Learning multi-modal distributions of pedestrian trajectories with gans,'' in \emph{Proc. IEEE/CVF Conf. Comput. Vis. Pattern Recognit. Workshops}, 2019, pp. 0--0.

\bibitem{chen2024wtgcn}
W.~Chen, H.~Sang, J.~Wang, and Z.~Zhao, ``Wtgcn: wavelet transform graph convolution network for pedestrian trajectory prediction,'' \emph{Int. J. Mach. Learn. Cybern.}, pp. 1--18, 2024.

\bibitem{CHEN2025104862}
------, ``Iggcn: Individual-guided graph convolution network for pedestrian trajectory prediction,'' \emph{Digital Signal Processing}, vol. 156, p. 104862, 2025.

\bibitem{li2024MFAN}
J.~Li, L.~Yang, Y.~Chen, and Y.~Jin, ``Mfan: Mixing feature attention network for trajectory prediction,'' \emph{Pattern Recognition}, vol. 146, p. 109997, 2024.

\bibitem{gu2021efficiently}
A.~Gu, K.~Goel, and C.~R{\'e}, ``Efficiently modeling long sequences with structured state spaces,'' \emph{arXiv preprint arXiv:2111.00396}, 2021.

\bibitem{gu2023mamba}
A.~Gu and T.~Dao, ``Mamba: Linear-time sequence modeling with selective state spaces,'' \emph{arXiv preprint arXiv:2312.00752}, 2023.

\bibitem{Ronen2022DeepDPMDC}
M.~Ronen, S.~E. Finder, and O.~Freifeld, ``Deepdpm: Deep clustering with an unknown number of clusters,'' in \emph{Proc. IEEE/CVF Conf. Comput. Vis. Pattern Recognit.}, 2022, pp. 9861--9870.

\bibitem{xiao2022deepplugandplay}
A.~Xiao, H.~Chen, T.~Guo, Q.~Zhang, and Y.~Wang, ``Deep plug-and-play clustering with unknown number of clusters,'' \emph{Trans. Mach. Learn. Res.}, 2022.

\bibitem{lv2023ssagcn}
P.~Lv, W.~Wang, Y.~Wang, Y.~Zhang, M.~Xu, and C.~Xu, ``Ssagcn: Social soft attention graph convolution network for pedestrian trajectory prediction,'' \emph{IEEE Transactions on Neural Networks and Learning Systems}, vol.~35, no.~9, pp. 11\,989--12\,003, 2024.

\end{thebibliography}

\begin{IEEEbiography}
[{\includegraphics[width=1in,height=2in,clip,keepaspectratio]{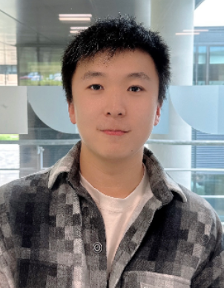}}]{Ruochen Li}
is a PhD candidate in the Department of Computer Science at Durham University, UK. His research interests include trajectory prediction, computer vision, time-series analysis, and graph convolutional networks.
\end{IEEEbiography}
\vspace{-4mm}

\begin{IEEEbiography}
[{\includegraphics[width=1in,height=1.25in,clip,keepaspectratio]{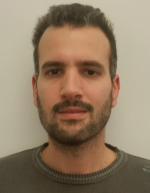}}]{Stamos Katsigiannis} (Member, IEEE) received the B.Sc. (Hons.) degree in informatics and telecommunications from the National and Kapodistrian University of Athens, Greece, in 2009,  the M.Sc. degree in computer science from the Athens University of Economics and Business, Greece, in 2011, and the Ph.D. degree in computer science (biomedical image and general-purpose video processing) from the National and Kapodistrian University of Athens, Greece, in 2016. He is currently an Associate Professor with the Department of Computer Science, Durham University, UK. He has participated in multiple national and international research projects and has authored and co-authored over 70 research publications. His research interests include machine learning, natural language processing, affective computing, image analysis, and image and video quality.
\end{IEEEbiography}
\vspace{-4mm}

\begin{IEEEbiography}
[{\includegraphics[width=1in,height=1.25in,clip,keepaspectratio]{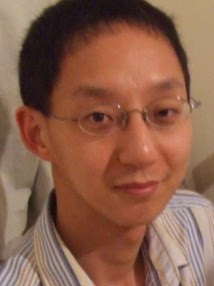}}]{Tae-Kyun (T-K) Kim} is a Professor and the director of Computer Vision and Learning Lab at School of Computing, KAIST, since 2020, and has been a Reader at Imperial College London, UK in 2010-2024. He obtained his PhD from Univ. of Cambridge in 2008 and Junior Research Fellowship (governing body) of Sidney Sussex College, Univ. of Cambridge during 2007-2010. His BSc and MSc are from KAIST. His research interests lie in machine (deep) learning for 3D vision and generative AI, he has co-authored over 100 academic papers in top-tier conferences and journals in the field. He was the general chair of BMVC17 in London, and the program co-chair of BMVC23, Associate Editor of Pattern Recognition Journal, Image and Vision Computing Journal. He regularly serves as an Area Chair for the top-tier AI/vision conferences.
\end{IEEEbiography}

\vspace{-4mm}

\begin{IEEEbiography}
[{\includegraphics[width=1in,height=1.25in,clip,keepaspectratio]{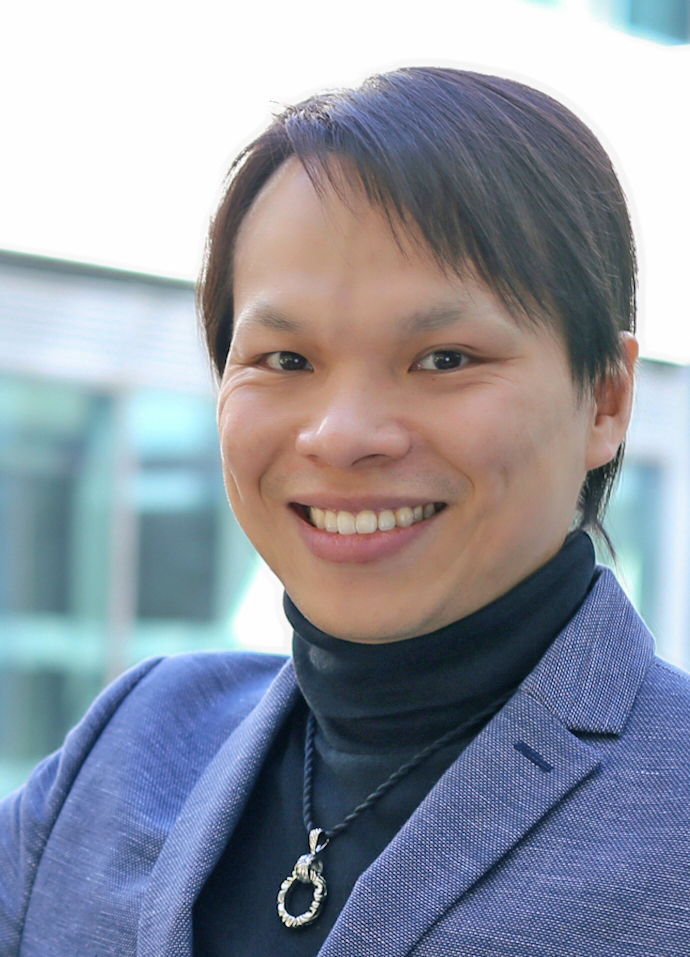}}]{Hubert P. H. Shum} (Senior Member, IEEE) is a Professor of Visual Computing and the Director of Research for the Department of Computer Science at Durham University, specialising in modelling spatio-temporal information with responsible AI. He is also a Co-Founder and the Co-Director of Durham University Space Research Centre. Before this, he was an Associate Professor at Northumbria University and a Postdoctoral Researcher at RIKEN Japan. He received his PhD degree from the University of Edinburgh. He chaired conferences such as Pacific Graphics, BMVC and SCA. He has authored over 200 research publications in the fields of Computer Vision, Computer Graphics and AI in Healthcare, underpinned by Responsible AI designs and algorithms.
\end{IEEEbiography}

\includepdf[pages=-]{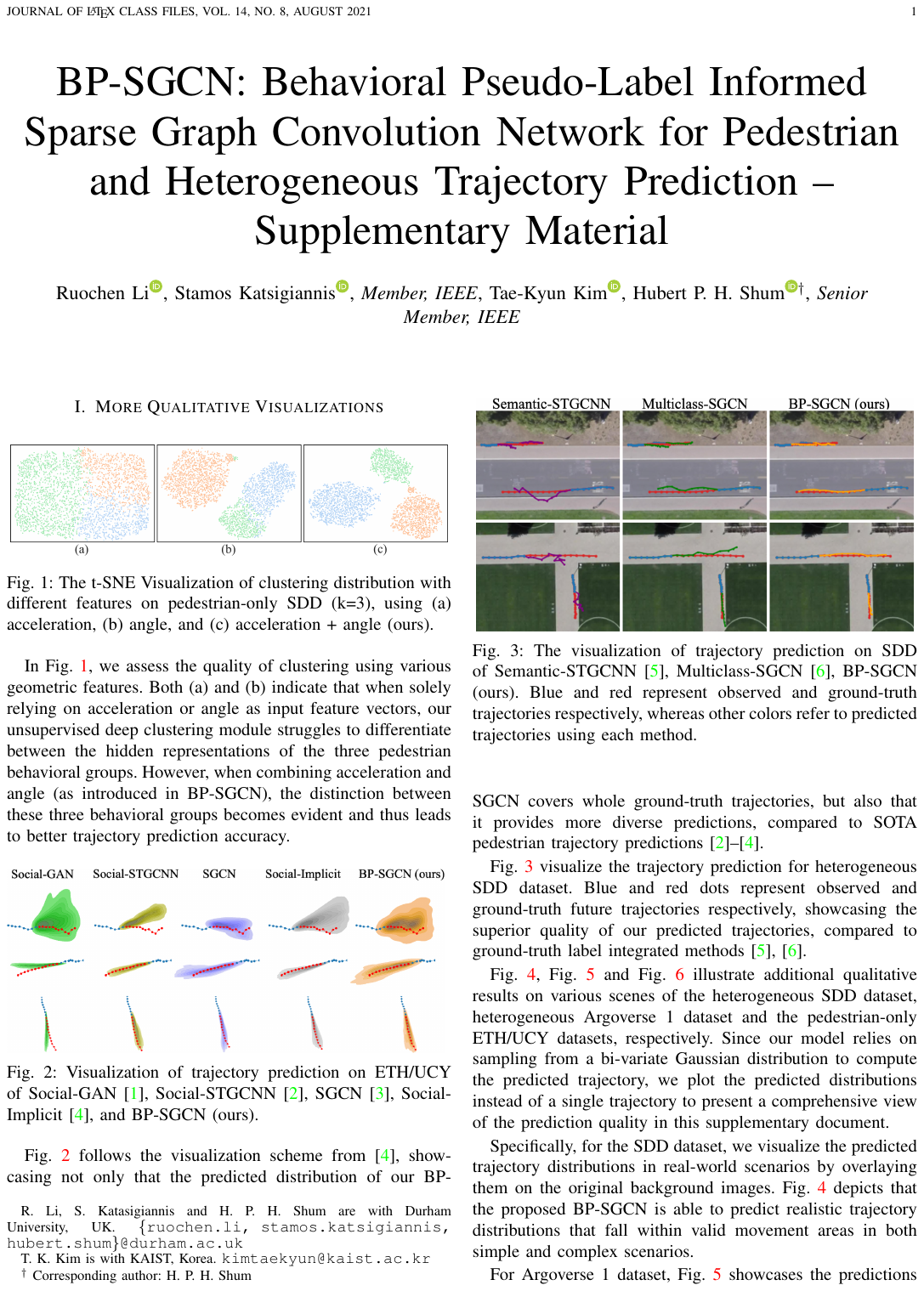}

\end{document}